\documentclass[final]{cvpr}

\usepackage{multirow}
\usepackage{times}
\usepackage{dsfont}
\usepackage{epsfig}
\usepackage{graphicx}
\usepackage{booktabs}
\usepackage{amsmath}
\usepackage{enumitem}
\usepackage[dvipsnames]{xcolor}
\usepackage{amssymb}
\usepackage[algo2e,ruled,vlined]{algorithm2e}
\usepackage{algorithm}
\usepackage{subcaption}
\usepackage{balance}
\usepackage{comment}
\usepackage{xcolor}

\usepackage[pagebackref=true,breaklinks=true,colorlinks,bookmarks=false]{hyperref}

\newcommand{\model}{\textsf{RUC}}

\newcommand{\cutparagraphup}{\vspace*{-0.15in}}

\def\cvprPaperID{3724} 
\def\confYear{CVPR 2021}

\begin{document}

\title{Improving Unsupervised Image Clustering With Robust Learning}

\author{Sungwon Park\thanks{Equal contribution to this work.}\textsuperscript{\rm ~~1,2}~~~Sungwon Han\footnotemark[1]\textsuperscript{\rm ~~1,2}~~~Sundong Kim\textsuperscript{\rm 2}~~~Danu Kim\textsuperscript{\rm 1,2} \\ Sungkyu Park\textsuperscript{\rm 2}~~~Seunghoon Hong\textsuperscript{\rm 1}~~~Meeyoung Cha\textsuperscript{\rm 2,1} \\

\hspace*{\fill}
\textsuperscript{\rm 1}School of Computing, KAIST 
\hspace{0.6cm} \textsuperscript{\rm 2}Data Science Group, Institute for Basic Science
\hspace*{\fill} \\
}

\maketitle
\thispagestyle{empty}
\pagestyle{empty}

\begin{abstract}
Unsupervised image clustering methods often introduce alternative objectives to indirectly train the model and are subject to faulty predictions and overconfident results. To overcome these challenges, the current research proposes an innovative model \model{} that is inspired by robust learning. \model{}'s novelty is at utilizing pseudo-labels of existing image clustering models as a noisy dataset that may include misclassified samples. Its retraining process can revise misaligned knowledge and alleviate the overconfidence problem in predictions. The model's flexible structure makes it possible to be used as an add-on module to other clustering methods and helps them achieve better performance on multiple datasets. Extensive experiments show that the proposed model can adjust the model confidence with better calibration and gain additional robustness against adversarial noise.
\end{abstract}
\section{Introduction}
Unsupervised clustering is a core task in computer vision that aims to identify each image's class membership without using any labels. Here, a class represents the group membership of images that share similar visual characteristics. Many studies have proposed deep learning-based algorithms that utilize distance in a feature space as the similarity metric to assign data points into classes~\cite{ding2004k,vincent2010stacked}.

\begin{figure}[t!]
    \centerline{
    \includegraphics[width=.98\columnwidth]{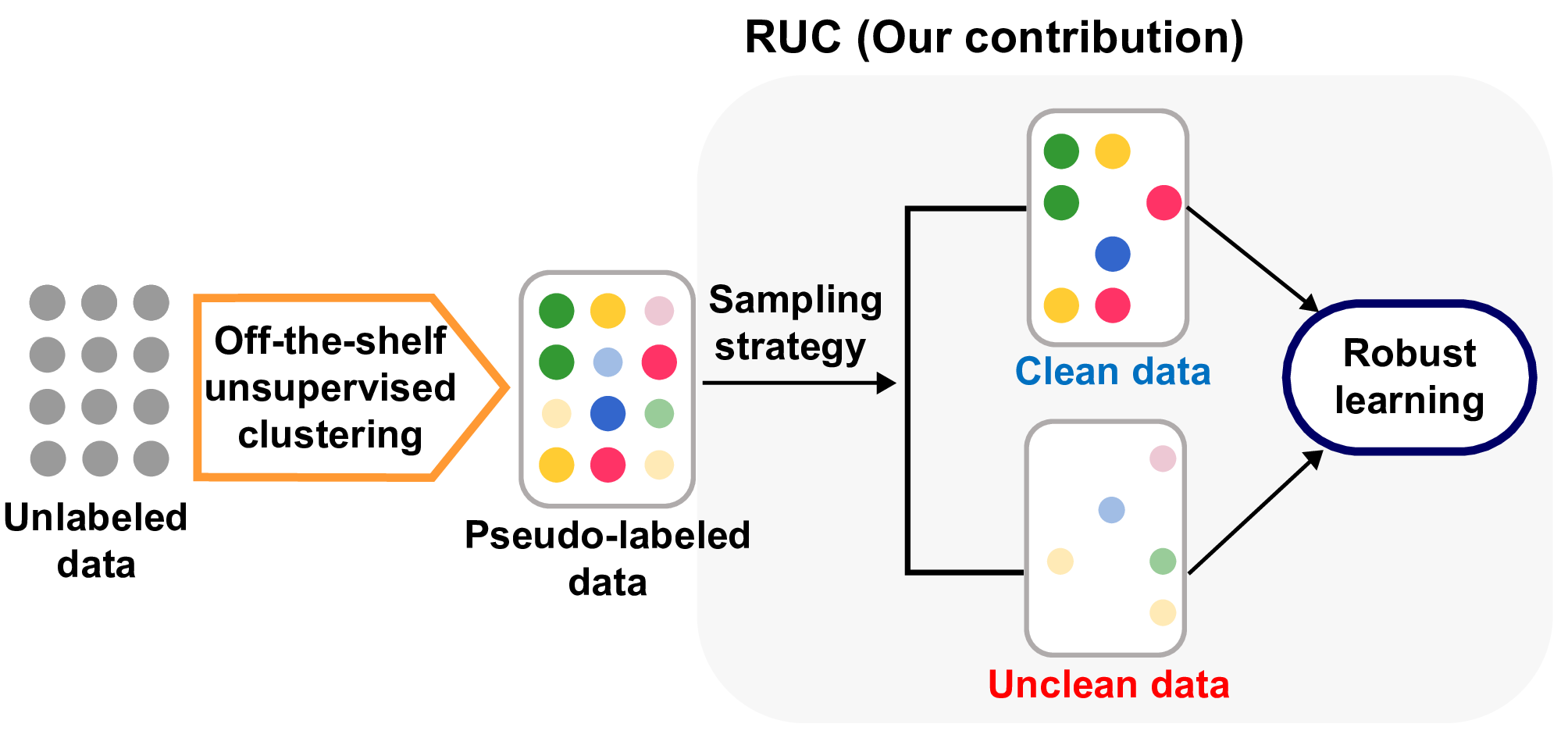}}
    \caption{Illustration for this work's basic concept: robust learning is used to separate clean data from unclean data using pseudo-labels from off-the-shelf unsupervised clustering algorithm.}
    \label{fig:intro}
\end{figure}

Training without ground-truth guidance, however, is prone to finding trivial solutions that are learned from low-level visual traits like colors and textures~\cite{ji2019invariant}. Several studies have introduced innovative ways to guide the model's training indirectly by setting alternative objectives. For example, Hu~\emph{et al.}~\cite{hu2017learning} proposed to maximize the mutual information between input and its hidden representations, and Ji~\emph{et al.}~\cite{ji2019invariant} proposed to learn invariant features against data augmentation. Entropy-based balancing has often been adopted to prevent degenerate solutions~\cite{hanmitigating,ji2019invariant, van2020scan}.

Nevertheless, these alternative objectives are bound to producing overconfident results, i.e., low-entropy predictions, due to the dense grouping among clusters. When uncertain samples are added to a wrong cluster at an early stage of training, the model gradually becomes overconfident in its later predictions as the noise from misclassification accumulates and degrades the overall performance.

This paper introduces a novel robust learning training method, \model{} (\textsf{R}obust learning for \textsf{U}nsupervised \textsf{C}lustering), that runs in conjunction with existing clustering models to alleviate the noise discussed above. Utilizing and treating the existing clustering model's results as a noisy dataset that may include wrong labels, \model{} updates the model's misaligned knowledge. Bringing insights from the literature, we filter out unclean samples and apply loss correction as in Fig.~\ref{fig:intro}. This process is assisted by label smoothing and co-training to reduce any wrong gradient signals from unclean labels. This retraining process with revised pseudo-labels further regularizes the model and prevents overconfident results.

\model{} comprises two key components: (1) extracting clean samples and (2) retraining with the refined dataset. We propose confidence-based, metric-based, and hybrid strategies to filter out misclassified pseudo-labels. The first strategy considers samples of high prediction confidence from the original clustering model as a clean set; it filters out low confidence samples. This strategy relies on the model's calibration performance. The second strategy utilizes similarity metrics from unsupervised embedding models to detect clean samples with non-parametric classifiers by checking whether the given instance shares the same labels with top $k$-nearest samples. The third strategy combines the above two and selects samples that are credible according to both strategies.

The next step is to retrain the clustering model with the sampled dataset. We use MixMatch~\cite{berthelot2019mixmatch}, a semi-supervised learning technique; which uses clean samples as labeled data and unclean samples as unlabeled data. We then adopt label smoothing to leverage strong denoising effects on the label noise~\cite{lukasik2020does} and block learning from overconfident samples~\cite{ji2019invariant,van2020scan}. Finally, a co-training architecture with two networks is used to mitigate noise accumulation from the unclean samples during training and increase performance.

We evaluate \model{} with rigorous experiments on datasets, including CIFAR-10, CIFAR-20, STL-10, and ImageNet-50. Combining \model{} to an existing clustering model outperforms the state-of-the-art results with the accuracy of 90.3\% in CIFAR-10, 54.3\% in CIFAR-20, 86.7\% in STL-10, and 78.5\% in ImageNet-50 dataset. \model{} also enhances the baseline model to be robust against adversarial noise. Our contributions are as follows: 
\begin{itemize}[noitemsep, topsep=2pt]
    \item The proposed algorithm \model{} aids existing unsupervised clustering models via retraining and avoiding overconfident predictions. 
    
    \item The unique retraining process of \model{} helps existing models boost performance. It achieves a 5.3pp increase for the STL-10 dataset when added to the state-of-the-art model (81.4\% to 86.7\%).
    
    \item The ablation study shows every component in \model{} is critical, including the three proposed strategies (i.e., confidence-based, metric-based, and hybrid) that excel in extracting clean samples from noisy pseudo-labels.
    
    \item The proposed training process is robust against adversarial noise and can adjust the model confidence with better calibrations.  
\end{itemize}

Implementation details of the model and codes are available at \url{https://github.com/deu30303/RUC}.

\section{Related Work}
\begin{figure*}[t!]
    \centerline{
    \includegraphics[width=1.9\columnwidth]{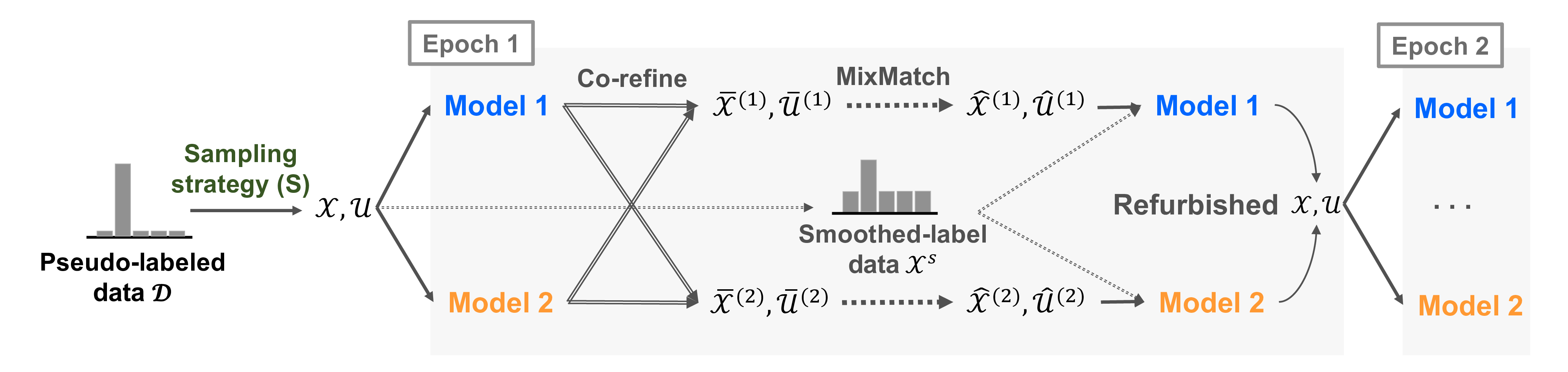}}
    \caption{Illustration of the proposed model. Our model first selects clean samples as a labeled dataset $\mathcal{X}$ and considers the remaining samples as an unlabeled dataset $\mathcal{U}$ (Section 3.1). Next, we train two networks $f_{\theta^{\tiny{(1)}}}$ and $f_{\theta^{\tiny{(2)}}}$ in a semi-supervised fashion (Section 3.2). In each epoch, the MixMatch algorithm, along with co-training and label smoothing, is applied for training. The clean set is updated via co-refurbishing for the next epoch.}
    \label{fig:mainmodel_fig}
\end{figure*}

\subsection{Unsupervised Image Clustering}
The main objective of clustering is to group the data points into distinct classes of similar traits~\cite{jain1999data}. Most real-world problems deal with high dimensional data (e.g., images), and thereby, setting a concrete notion of similarity while extracting low-dimensional features becomes key components for setting appropriate standards for grouping~\cite{xie2016unsupervised}. Likewise, unsupervised clustering is a line of research aiming to tackle both dimensionality reduction and boundary identification over the learned similarity metric~\cite{hanmitigating}. Existing research can be categorized into \emph{sequential}, \emph{joint}, and \emph{multi-step refinement approach}.
\cutparagraphup
\paragraph{Sequential approach.} Sequential approach extracts features, then sequentially applies the conventional distance or density-based clustering algorithm for class assignments. For example, Ding~\emph{et al.}~\cite{ding2004k} use principal component analysis to extract low-dimensional features and then apply $k$-means clustering to assign classes. For feature extraction, autoencoder structures are often used to extract latent features before grouping, types of autoencoder include stacked~\cite{vincent2010stacked}, boolean~\cite{baldi2012autoencoders}, or variational autoencoder~\cite{kingma2013auto}. However, these models tend to produce features with little separation among clusters due to the lack of knowledge on subsequent assignment processes. 
\cutparagraphup
\paragraph{Joint approach.} The joint approach's characteristic is to use an end-to-end pipeline that concurrently performs feature extraction and class assignment. An example is Yang~\emph{et al.}~\cite{yang2017towards}, which adopt the concept of clustering loss to guarantee enough separations among clusters. End-to-end CNN pipelines are used widely to iteratively identify clusters while refining extracted features~\cite{caron2018deep,chang2017deep,xie2016unsupervised}. Recent studies have shown that a mutual information-based objective is an effective measure to improve classification accuracy~\cite{hu2017learning,ji2019invariant}. Nonetheless, those models still bear the problem of generating unintended solutions that depend on trivial low-level features from random initialization~\cite{hanmitigating}. 
\cutparagraphup
\paragraph{Multi-step refinement approach.} To mitigate the unintended trivial solutions, recent approaches leverage the power of unsupervised embedding learning models to provide better initialization for downstream clustering tasks~\cite{chen2020simple,wu2018unsupervised,2020superand}. These methods generate feature representations to gather data points with similar visual traits and push away the rest in an embedding space. With the initialization, clustering results are elaborated in a refinement step, bringing significant gain in its class assignment quality~\cite{hanmitigating,van2020scan}. In particular, SCAN~\cite{van2020scan} first obtains high-level feature representations by feature similarity then clusters those representations by nearest neighbors, and this model has shown remarkable performance on unsupervised clustering.

\cutparagraphup
\paragraph{Add-on modules to improve unsupervised clustering.}
The proposed retraining process with sample selection strategy improves off-the-shelf unsupervised clustering algorithms (e.g., sequential, joint, multi-step refinement) by acting as an add-on module. Our module's main objective is to revise the misaligned knowledge of trained clustering models via label cleansing and retraining with the refined labels. This method has not been well investigated before but has begun to be proposed recently. Gupta~\emph{et al.}~\cite{gupta2020unsupervised} show that semi-supervised retraining improves unsupervised clustering. They draw a graph where data samples are nodes, and the confidence from ensemble models between the samples is an edge. Then, a dense sub-graph is considered as a clean set. \looseness=-1

The main difference between Gupta~\emph{et al.} and ours is in how we treat the pseudo-labels obtained by the clustering. Gupta~\emph{et al.} treats the pseudo-label as a ground-truth for semi-supervised learning, which produces sub-optimal result if the pseudo-label is noisy (i.e., \textit{memorization}). In contrast, we introduce the robust learning concept of label smoothing and co-training to mitigate the memorization of noisy samples, which leads to substantial improvements in the calibration and clustering performance.

\subsection{Robust Learning With Label Noise}
A widely used setting for robust learning is where an adversary has deliberately corrupted the labels, which otherwise arise from some clean
distribution~\cite{natarajan2013learning, song2020learning}. According to the literature, deep networks easily overfit to the label noise during training and get a low generalization power~\cite{liu2020early}. In this light, models that prevent overfitting in a noise label environment have been studied. 
\cutparagraphup
\paragraph{Loss correction.} The first representative line of work is a loss correction, which relabels unclean samples explicitly or implicitly. For example, Patrini~\emph{et al.}~\cite{patrini2017making} estimate the label transition probability matrix to correct the loss and retrain the model. To estimate the transition matrix more accurately, the gold loss correction approach~\cite{hendrycks2018using} is proposed to utilize trusted labels as additional information.
\cutparagraphup
\paragraph{Loss reweighting.} The next line of work is loss reweighting, which aims to give a smaller weight to the loss of unclean samples so that model can reduce the negative effect of label noise during training. One work computes the importance as an approximated ratio of two data distributions; clean and unclean~\cite{wang2017multiclass}. On the other hand, the active bias approach~\cite{chang2017active} calculates the inconsistency of predictions during training and assigns a weight to penalize unclean data. \looseness=-1
\cutparagraphup
\paragraph{Sample selection.} Relabeling the misclassified samples may cause a false correction. In this context, recent works introduce a sample selection approach that filters out misclassified samples and only selects clean data for training~\cite{lyu2019curriculum,malach2017decoupling}. Notably, the small loss trick, which regards the sample with small training loss as clean, effectively separates true- and false-labeled data points~\cite{arpit2017closer,jiang2018mentornet,liu2020early}. Also, recent studies suggest diverse ways to lead additional performance by maintaining two networks to avoid accumulating sampling bias~\cite{han2018co,yu2019does}, adopting refurbishment of false-labeled samples~\cite{song2019selfie}, or using a semi-supervised approach to utilize false-labeled sample maximally~\cite{li2020dividemix}. Our model advances some of these sample selection approaches to filter out unclean samples out of clustering results and utilize clean samples only during retraining.
\section{Method}
\text{\model} is an add-on method that can be used in conjunction with the existing unsupervised clustering methods to refine mispredictions. Its key idea is at utilizing the initial clustering results as \emph{noisy} pseudo-labels and learning to refine them with a mild clustering assumption~\cite{Engelen2020survey} and techniques from the robust learning~\cite{li2020dividemix,lukasik2020does}. 

Figure~\ref{fig:mainmodel_fig} and Algorithm~\ref{algo:overall} illustrate the overall pipeline of the proposed algorithm. Given the initial pseudo-labels, we first divide the training data into the two disjoint sets: clean and unclean (Section~\ref{sec:cleaning}). Then treating these sets each as labeled and unlabeled data, we train a classifier in a semi-supervised manner while refurbishing the labeled and unlabeled data (Section~\ref{sec:semisup}). We guide the semi-supervised class assignment with robust learning techniques, such as co-training and label smoothing, to account for inherent label noises. These techniques are useful in handling label noises and calibrating the model's prediction score. Below we describe the model in details.

\begin{algorithm*}[t!]
\DontPrintSemicolon
\SetAlgoLined
\SetNoFillComment 
\begin{flushleft}
 \textbf{Input:} Sampling strategy $\mathcal{S}$, training dataset with pseudo-labels $\mathcal{D}$, two networks $f_{\theta^{\tiny{(1)}}}$,  $f_{\theta^{\tiny{(2)}}}$, sharpening temperature $T$, number of augmentations $M$, unsupervised loss weight $\lambda_{\mathcal{U}}$, refurbish threshold $\tau_{2}$, weak- and strong augmentation $\phi_{a}$, $\phi_{A}$ \smallskip
 
\small{
 \tcc{Divide the dataset $\mathcal{D}$ into clean and noisy set using a sampling strategy}
 $\mathcal{X}, \mathcal{U} = \mathcal{S}(\mathcal{D})$  \space\space (i.e. $\mathcal{X} = \{({\mathbf{x}_{b}}, \mathbf{y}_{b}): b \in (1, ... , B)\}$,   $\mathcal{U} = \{\mathbf{u}_{b}: b \in (1, ... , B)\}$)

\For{$k \in \{1,2\}$}
{
 \tcc{Train the two networks $f_{\theta^{\tiny{(1)}}}$ and $f_{\theta^{\tiny{(2)}}}$ iteratively}
\For{$b  \in \{1,..,B\}$}
{
$\Tilde{\mathbf{y}}_{b} = (1 - \epsilon) \cdot \mathbf{y}_{b} + {\epsilon \over (C - 1)} \cdot (\mathbf{1} - \mathbf{y}_{b}) $ \tcp*{Inject uniform noise into all classes (label smoothing)}
 \For{$m  \in \{1,..,M\}$}
 {
    $\mathbf{x}_{b,m}, \mathbf{u}_{b,m} = \phi_{a}(\mathbf{x}_{b}), \phi_{a}(\mathbf{u}_{b})$  \tcp*{Perform weak augmentation $M$ times}
 }
  $\bar{\mathbf{y}}_{b} = (1 - w_{b}^{(c)}) \cdot \mathbf{y}_{b} + w_{b}^{
  (c)}\cdot f_{\theta^{(c)}}(\mathbf{x}_{b})$ \tcp*{Refine the labels ($(c)$ denotes the counter network)} 
  $\bar{\mathbf{y}}_{b} = \text{Sharpen}(\bar{\mathbf{y}}_{b} ,  T)$ \tcp*{Apply sharpening to the refined label}
   
  $\bar{\mathbf{q}}_{b} = {1 \over 2M} \sum_{m} (f_{\theta^{(1)}}(\mathbf{u}_{b,m}) + f_{\theta^{(2)}}(\mathbf{u}_{b,m}))$ \tcp*{Ensemble both networks' predictions to guess labels}
  $\bar{\mathbf{q}}_{b} = \text{Sharpen}(\bar{\mathbf{q}_{b}} ,  T)$ \tcp*{Apply sharpening to the guessed labels}
}

$\mathcal{X}^{s} = \{ (\phi_{A}(\mathbf{x}_{b}), \Tilde{\mathbf{y}}_{b}); b \in \{1,...,B\}\} $  \tcp*{Strongly augmented samples with smoothed labels}
$\bar{\mathcal{X}}^{(k)} = \{ (\mathbf{x}_{b}, \bar{\mathbf{y}}_{b}); b \in \{1,...,B\}\} $ \tcp*{Co-refined labeled samples}
$\bar{\mathcal{U}}^{(k)} = \{ (\mathbf{u}_{b}, \bar{\mathbf{q}}_{b}); b \in \{1,...,B\}\} $ \tcp*{Co-refined unlabeled samples}
$\hat{\mathcal{X}}^{(k)}, \hat{\mathcal{U}}^{(k)}$ = MixMatch($\bar{\mathcal{X}}^{(k)},\bar{\mathcal{U}}^{(k)}$)  \tcp*{Apply MixMatch}
$\mathcal{L}_{t} =
\mathcal{L}_{\mathcal{X}^s} + \mathcal{L}_{\mathcal{\hat{X}}} +  \mathcal{\lambda}_{\mathcal{U}}\mathcal{L}_{\mathcal{\hat{U}}}$  \tcp*{Calculate the total loss}
$\mathcal{X} \leftarrow \mathcal{X} \cup \text{Co-Refurbish}(\mathcal{U}, f_{\theta^{(k)}}, \tau_{2}) $  \tcp*{Refurbish noisy samples to clean samples (Eq.~\eqref{eq:refurb1},~\eqref{eq:refurb2})}
$\theta^{(k)} \leftarrow SGD(L_{t}, \theta^{(k)})$ \tcp*{Update network parameters}}} \vspace{-3mm}
\end{flushleft}
\caption{Robust learning algorithm using unsupervised clustering pseudo-label.}
\label{algo:overall}
\end{algorithm*}

\subsection{Extracting Clean Samples}
\label{sec:cleaning}
Let $\mathcal{D} = \{({\mathbf{x}_{i}}, \mathbf{y}_{i})\}_{i=1}^N$ denote the training data, where $\mathbf{x}_i$ is an image and $\mathbf{y}_i=g_\phi(\mathbf{x}_i)$ is a pseudo-label from an unsupervised classifier $g_\phi$.
The model first divides the pseudo-labels into two disjoint sets as $\mathcal{D}=\mathcal{X}\cup\mathcal{U}$ with a specified sampling strategy. 
We consider $\mathcal{X}$ as clean, whose pseudo-labels are moderately credible and thus can be used as a labeled dataset $(\mathbf{x},\mathbf{y})\in\mathcal{X}$ for refinement. In contrast, we consider $\mathcal{U}$ as unclean, whose labels we discard $\mathbf{u}\in\mathcal{U}$.
Designing an accurate sampling strategy is not straightforward, as there is no ground-truth to validate the pseudo-labels directly. Inspired by robust learning's clean set selection strategy, we explore three different approaches: (1) confidence-based, (2) metric-based, and (3) hybrid. \looseness=-1

\vspace{-1mm}
\cutparagraphup
\paragraph{Confidence-based strategy.} 
This approach selects clean samples based on the confidence score of the unsupervised classifier.
Given a training sample $(\mathbf{x},\mathbf{y})\in\mathcal{D}$, we consider the pseudo-label $\mathbf{y}$ is credible if $\max(\mathbf{y})>\tau_{1}$, and add it to the clean set $\mathcal{X}$. Otherwise, it is assigned to $\mathcal{U}$.
This is motivated by the observation that the unsupervised classifier tends to generate overconfident predictions; thus, we trust only the most typical examples from each class while ignoring the rest. The threshold $\tau_{1}$ is set substantially high to eliminate as many uncertain samples as possible.

\cutparagraphup
\paragraph{Metric-based strategy.} 
The limitation of the above approach is that the selection strategy still entirely depends on the unsupervised classifier. The metric-based approach leverages an additional embedding network $h_\psi$ learned in an unsupervised manner (e.g., SimCLR~\cite{chen2020simple}) and measures the credibility of the pseudo-label based on how well it coincides to the classification results using $h_\psi$.
For each $(\mathbf{x},\mathbf{y})\in\mathcal{D}$, we compute its embedding $h_\psi(\mathbf{x})$, and apply the non-parameteric classifier based on \emph{k}-Nearest Neighbor (\emph{k}-NN) by $\mathbf{y}'=k\text{-NN}(h_\psi(\mathbf{x}))$.
We consider that the pseudo-label is credible if $\text{argmax}(\mathbf{y})=\text{argmax}(\mathbf{y}')$ and add it to the clean set $\mathcal{X}$. Otherwise, it is assigned to the unclean set $\mathcal{U}$.

\cutparagraphup
\paragraph{Hybrid strategy.} 
This approach will add a sample to the clean set $\mathcal{X}$ only if it is considered credible by both the confidence-based and metric-based strategies. All other samples are added to $\mathcal{U}$. 
 
\subsection{Retraining via Robust Learning}
\label{sec:semisup}
Given the clean set $\mathcal{X}$ and the unclean set $\mathcal{U}$, our next step aims to train the refined classifier $f_\theta$ that revises incorrect predictions of the initial unsupervised classifier.

\cutparagraphup
\paragraph{Vanilla semi-supervised learning.} 
A naive baseline is to consider $\mathcal{X}$ as labeled data and $\mathcal{U}$ as unlabeled data each to train a classifier $f_\theta$ using semi-supervised learning techniques.
We utilize MixMatch~\cite{berthelot2019mixmatch} as such a baseline, which is a semi-supervised algorithm that estimates low-entropy mixed labels from unlabeled examples using MixUp augmentation~\cite{zhang2017mixup}.\footnote{Note that our method is not dependent on the particular choice of the semi-supervised learning method and can incorporate the others.}
For unsupervised clustering, MixUp can bring additional resistance against noisy labels since a large amount of extra virtual examples from MixUp interpolation makes memorization hard to achieve~\cite{li2020dividemix, zhang2017mixup}. 
Specifically, given a two paired data $(\mathbf{x}_1, \mathbf{y}_1)$ and $(\mathbf{x}_2, \mathbf{y}_2)$ sampled from either labeled or unlabeled data, it augments the data using the following operations. \smallskip
\begin{align}
    \lambda &\sim Beta(\alpha, \alpha) \\
    \lambda' &= max(\lambda, 1-\lambda) \\
    \mathbf{x}' &= \lambda' \mathbf{x}_1 + (1 - \lambda') \mathbf{x}_2 \\
    \mathbf{y}' &= \lambda' \mathbf{y}_1 + (1 - \lambda') \mathbf{y}_2.
    \label{eqn:mixup}
\end{align}
In the case of unlabeled data $\mathbf{u}\in\mathcal{U}$, MixMatch is employed such that a surrogate label $\mathbf{y}=\mathbf{q}$ is obtained by averaging the model's predictions over multiple augmentations after sharpening~\cite{berthelot2019mixmatch}.
Later, we will show that using the labels $\mathbf{y}$ and $\mathbf{q}$ directly in semi-supervised learning leads to a suboptimal solution and discuss how to improve its robustness.

For $\hat{\mathcal{X}}$ and $\hat{\mathcal{U}}$ after MixMatch (Eq.~\eqref{eq:mixmatch_vanilla}), a vanilla semi-supervised learning model trains with two separate losses: the cross-entropy loss for the labeled set $\hat{\mathcal{X}}$ (Eq.~\eqref{eq:m_loss}), and the consistency regularization for the unlabeled set $\hat{\mathcal{U}}$ (Eq.~\eqref{eq:u_loss}). $H(p, q)$ denotes the cross-entropy between $p$ and $q$.
\begin{align}
\hat{\mathcal{X}}&, \hat{\mathcal{U}} = \text{MixMatch}(\mathcal{X}, \mathcal{U}) \label{eq:mixmatch_vanilla} \\
\mathcal{L}_{\hat{\mathcal{X}}} &= {1 \over |\hat{\mathcal{X}}|} \sum_{\hat{\mathbf{x}}, \hat{\mathbf{y}} \in \hat{\mathcal{X}}} H(\hat{\mathbf{y}}, f_{\theta}(\hat{\mathbf{x}}))  
\label{eq:m_loss} \\
\mathcal{L}_{\hat{\mathcal{U}}} &= {1 \over |\hat{\mathcal{U}}|} \sum_{\hat{\mathbf{u}}, \hat{\mathbf{q}} \in \hat{\mathcal{U}}} ||\hat{\mathbf{q}} -  f_{\theta}(\hat{\mathbf{u}}) ||_{2}^2  
\label{eq:u_loss}
\end{align}

\cutparagraphup
\paragraph{Label smoothing.} To regularize our model from being overconfident to noisy predictions, we apply label smoothing along with vanilla semi-supervised learning.
Label smoothing prescribes soft labels by adding uniform noise and improves the calibration in predictions~\cite{lukasik2020does}.
Given a labeled sample with its corresponding label $(\mathbf{x},\mathbf{y})\in\mathcal{X}$, we inject uniform noise into all classes as follows: 
\begin{equation}
    \Tilde{\mathbf{y}} = (1 - \epsilon) \cdot \mathbf{y} + {\epsilon \over (C - 1)} \cdot (\mathbf{1} - \mathbf{y})
    \label{eq:label_smoothing}
\end{equation} 
where C is the number of class and $\epsilon\sim\text{Uniform}(0,1)$ is the noise. 
We compute cross-entropy using the soft label $\Tilde{\mathbf{y}}$ and the predicted label of the strongly augmented sample $\phi_{A}(\mathbf{x})$ via RandAugment~\cite{cubuk2020randaugment}. We find that strong augmentations minimize the memorization from noise samples. \looseness=-1 
\begin{equation}
\mathcal{L}_{\mathcal{X}^s} = {1 \over |\mathcal{X}|} \sum_{\mathbf{x}, \Tilde{\mathbf{y}} \in \mathcal{X}} H(\Tilde{\mathbf{y}}, f_{\theta}(\phi_{A}(\mathbf{x}))) 
\label{eq:s_loss}
\end{equation}
Our final objective for training can be written as:
\begin{equation}
\mathcal{L}(\theta;\mathcal{X}, \hat{\mathcal{X}}, \hat{\mathcal{U}}) =
\mathcal{L}_{\mathcal{X}^s} + \mathcal{L}_{\hat{\mathcal{X}}} +  \mathcal{\lambda}_{\mathcal{U}}\mathcal{L}_{\hat{\mathcal{U}}},
\label{eq:t_loss}
\end{equation}
where $\lambda_{\mathcal{U}}$ is a hyper-parameter to control the effect of the unsupervised loss in MixMatch.

\cutparagraphup 
\paragraph{Co-training.} 
Maintaining a single network for learning has a vulnerability of overfitting to incorrect pseudo-labels since the initial error from the network is transferred back again, and thereby, accumulated~\cite{han2018co}. To avoid this fallacy, we additionally introduce a co-training module where the two networks $f_{\theta^{(1)}}, f_{\theta^{(2)}}$ are trained in parallel and exchange their guesses for teaching each other by adding a co-refinement step on top of MixMatch. \looseness=-1

Co-refinement is a label refinement process that aims to produce reliable labels by incorporating both networks' predictions. Following the previous literature~\cite{li2020dividemix}, we apply co-refinement both on the label set $\mathcal{X}$ and the unlabeled set $\mathcal{U}$ for each network. Here, we explain the co-refinement process from the perspective of $f_{\theta^{(1)}}$. 
For the labeled data point $\mathbf{x}$, we calculate the linear sum between the original label $\mathbf{y}$ in $\mathcal{X}$ and the prediction from the counter network $f_{\theta^{(2)}}$ (Eq.~\eqref{eq:label_refine}) and apply sharpening on the result to generate the refined label $\bar{\mathbf{y}}$ (Eq.~\eqref{eq:sharp}).
\begin{align}
    \bar{\mathbf{y}} &= (1 - w^{(2)}) \cdot \mathbf{y} + w^{(2)}\cdot f_{\theta^{(2)}}(\mathbf{x})
    \label{eq:label_refine} \\
    \bar{\mathbf{y}} &= \text{Sharpen}(\bar{\mathbf{y}} ,  T),
    \label{eq:sharp}
\end{align} 
where $w^{(2)}$ is the counter network's confidence value of $\mathbf{x}$, and $T$ is the sharpening temperature. For the unlabeled set $\mathcal{U}$, we apply an ensemble of both networks' predictions to guess the pseudo-label $\bar{\mathbf{q}}$ of data sample $\mathbf{u}$ as follows:
\begin{align}
 \bar{\mathbf{q}} &= {1 \over 2M} \sum_{m} (f_{\theta^{(1)}}(\mathbf{u}_{m}) + f_{\theta^{(2)}}(\mathbf{u}_{m})) 
  \label{eq:label_guess}\\
 \bar{\mathbf{q}} &= \text{Sharpen}(\bar{\mathbf{q}} ,  T),
 \label{eq:sharp2}
\end{align}
where $\mathbf{u}_{m}$ is $m$-th weak augmentation of $\mathbf{u}$.

In place of $\mathcal{X}$ and ~$\mathcal{U}$, co-refinement produces the refined dataset 
$(\mathbf{x}, \bar{\mathbf{y}})\in\mathcal{\bar{X}}^{(1)}$, and $(\mathbf{u},\bar{\mathbf{q}})\in\bar{\mathcal{U}}^{(1)}$ through Eq.~\eqref{eq:label_refine} to \eqref{eq:sharp2}. We utilize those datasets as an input for MixMatch, and the model is eventually optimized as follows: 
\begin{align}
    \mathcal{\bar{X}}^{(1)}, \mathcal{\bar{U}}^{(1)} &= \text{Co-refinement}(\mathcal{X}, \mathcal{U}, \theta^{(1)}, \theta^{(2)}) \\
    \mathcal{\hat{X}}^{(1)}, \mathcal{\hat{U}}^{(1)} &= \text{MixMatch}(\mathcal{\bar{X}}^{(1)}, \mathcal{\bar{U}}^{(1)})\\
    \theta^{(1)} &\gets \arg\min_{\theta^{(1)}} \mathcal{L}(\theta^{(1)};\mathcal{X}, \hat{\mathcal{X}}^{(1)}, \hat{\mathcal{U}}^{(1)}),
\end{align}
where $\mathcal{L}$ is the loss defined in Eq.~\eqref{eq:t_loss}. 
This process is also conducted for $f_{\theta^{(2)}}$ in the same manner. 

\cutparagraphup
\paragraph{Co-refurbishing.}
Lastly, we refurbish the noise samples at the end of every epoch to deliver the extra clean samples across the training process. 
If at least one of the networks' confidence on the given unclean sample $\mathbf{u}\in\mathcal{U}$ is over the threshold $\tau_{2}$, the corresponding sample's label is updated with the network's prediction $\mathbf{p}$. The updated sample is then regarded as clean and appended to the labeled set $\mathcal{X}$.
\begin{align}
\hspace{-2mm}\mathbf{p} &= f_{\theta^{(k)}} (\mathbf{u}) \text{, where } k = \underset{k'}{\arg\max} (\max( f_{\theta^{(k')}} (\mathbf{u}) ))\label{eq:refurb1} \looseness=-1 \\
\hspace{-2mm}\mathcal{X} &\leftarrow \mathcal{X} \cup \{ (\mathbf{u}, \mathds{1}_\mathbf{p}) | \max (\mathbf{p}) > \tau_2 \}, \label{eq:refurb2}
\end{align}
where $\mathds{1}_\mathbf{p}$ is a one-hot vector of $\mathbf{p}$ whose  $i$-th element is 1, considering $i = \arg\max(\mathbf{p})$.

\section{Experiments}
For evaluation, we first compared the performance of our model against other baselines over multiple datasets. Then, we examined each component's contribution to performance improvement. Lastly, we investigated how \model{}~helps improve existing clustering models in terms of their confidence calibration and robustness against adversarial attacks. \looseness=-1
\subsection{Unsupervised Image Clustering Task}

\paragraph{Settings.} Four benchmark datasets were used. The first two are CIFAR-10 and CIFAR-100, which contain 60,000 images of 32x32 pixels. For CIFAR-100, we utilized 20 superclasses following previous works~\cite{van2020scan}. The next is STL-10, containing 100,000 unlabeled images and 13,000 labeled images of 96x96 pixels. For the clustering problem, only 13,000 labeled images were used. Lastly, we test the model with the large-scale ImageNet-50 dataset, containing 65,000 images of 256x256 pixels.

Our model employed the ResNet18~\cite{he2016deep} architecture following other baselines~\cite{hanmitigating,ji2019invariant,van2020scan} and the model was trained for 200 epochs. Initial confidence threshold $\tau_{1}$ was set as 0.99, and the number of neighbors $k$ to divide the clean and noise samples was set to 100. The threshold $\tau_{2}$ for refurbishing started from 0.9 and increased by 0.02 in every 40 epochs. The label smoothing parameter $\epsilon$ was set to 0.5. For evaluating the class assignment, the Hungarian method~\cite{kuhn1955hungarian} was used to map the best bijection permutation between the predictions and ground-truth.
\cutparagraphup
\paragraph{Result.} Table~\ref{Tab:classification} shows the overall performance of clustering algorithms over three datasets: CIFAR-10, CIFAR-20, and STL-10. For these datasets, the proposed model \model{}, when applied to the SCAN~\cite{van2020scan} algorithm, outperforms all other baselines. Particularly for STL-10, the combined model shows a substantial improvement of 5.3~pp. 
Table~\ref{Tab:ImageNet} reports ImageNet-50 result on the confidence based sampling strategy, which demonstrates \model{}'s 
applicability to large-scale dataset.
Furthermore, \model{} achieves consistent performance gain over another clustering model, TSUC~\cite{hanmitigating}. These results confirm that our model can be successfully applied to existing clustering algorithms and improve them. We also confirm that all three selection strategies (i.e., confidence-based, metric-based, and hybrid) bring considerable performance improvement. 

\begin{table}[t!]
\scalebox{0.83}{
\begin{tabular}{l|ccc}
\toprule
Method              & CIFAR-10 & CIFAR-20 & STL-10  \\ \midrule
\emph{k}-means~\cite{wang2014optimized}             & 22.9     & 13.0     & 19.2    \\
Spectral clustering~\cite{zelnik2005self}  & 24.7     & 13.6     & 15.9    \\
Triplets~\cite{schultz2004learning}          & 20.5     & 9.9      & 24.4    \\
Autoencoder (AE)~\cite{bengio2007greedy}                   & 31.4     & 16.5     & 30.3   \\
Variational Bayes AE~\cite{kingma2013auto} & 29.1     & 15.2     & 28.2    \\
GAN  ~\cite{radford2015unsupervised}           & 31.5     & 15.1     & 29.8    \\
JULE ~\cite{yang2016joint}           & 27.2     & 13.7     & 27.7    \\
DEC  ~\cite{xie2016unsupervised}           & 30.1     & 18.5     & 35.9    \\
DAC ~\cite{chang2017deep}            & 52.2     & 23.8     & 47.0    \\
DeepCluster ~\cite{caron2018deep}    & 37.4     & 18.9     & 33.4    \\
ADC~\cite{haeusser2018associative}             & 32.5     & 16.0     & 53.0    \\
IIC~\cite{ji2019invariant}            & 61.7     & 25.7     & 49.9    \\ 
TSUC$\dagger$~\cite{hanmitigating}                            & 80.2 & 35.5 &  62.0\\ 
SCAN$\dagger$~\cite{van2020scan}                            & 88.7 & 50.6 & 81.4 \\ \midrule
TSUC + \model{}~(Confidence) & 81.8 / 82.5 & 39.6 / 40.6 &  65.1 / 65.5 \\
TSUC + \model{}~(Metric) & 82.5 / 82.9  & 39.5 / 40.4  & 66.3 / 66.6\\
TSUC + \model{}~(Hybrid) &  82.1 / 82.8   &  39.5 / 40.6  & 66.0 / 66.8\\\midrule
SCAN + \model{}~(Confidence) &   \textbf{90.3} / 90.3   & 53.3 / 53.5     &\textbf{86.7}  / 86.8 \\
SCAN + \model{}~(Metric) &   89.5 / 89.5  &   53.9 / 53.9  &  84.7  / 85.1\\ 
SCAN + \model{}~(Hybrid) &  90.1/ 90.1   & \textbf{54.3} / 54.5  & 86.6 / 86.7\\\bottomrule
\end{tabular}}
\caption{Performance improvement with \model{}~(accuracy presented in percent). Baseline results are excerpted from~\cite{hanmitigating, van2020scan} and we report the last/best accuracy. $\dagger$Results obtained from our experiments with official code.}  
\label{Tab:classification}
\end{table}

\begin{table}[!h]
\centering
\scalebox{0.8}{
\begin{tabular}{l|cc}
\toprule
Method & SCAN (Best) & SCAN + \model{} (Last / Best accuracy) \\ \midrule
ImageNet-50 & 76.8 & \textbf{78.5} / 78.5 \\
\bottomrule 
\end{tabular}}
\caption{Performance comparison over ImageNet-50}
\label{Tab:ImageNet}
\end{table}

\begin{table}[!h]
\centering
\scalebox{0.9}{\begin{tabular}{l|cc}
\toprule
Setup & Last Acc & Best Acc            \\ \midrule
\model{} with all components     & \textbf{86.7}  & \textbf{86.8}\\
without co-training              &      86.2              & 86.4\\ 
without label smoothing   &       85.5  & 85.8 \\
with MixMatch only                  &     85.2      & 85.4\\
\bottomrule 
\end{tabular}}
\caption{Ablation results of the SCAN+\model{} on STL-10}
\label{Tab: Ablation}
\end{table}

\subsection{Component Analyses}

\begin{figure*}[!t]
\begin{minipage}[b]{0.98\linewidth}
\centering
    \begin{subfigure}[b]{0.22\textwidth}
      \includegraphics[width=\textwidth]{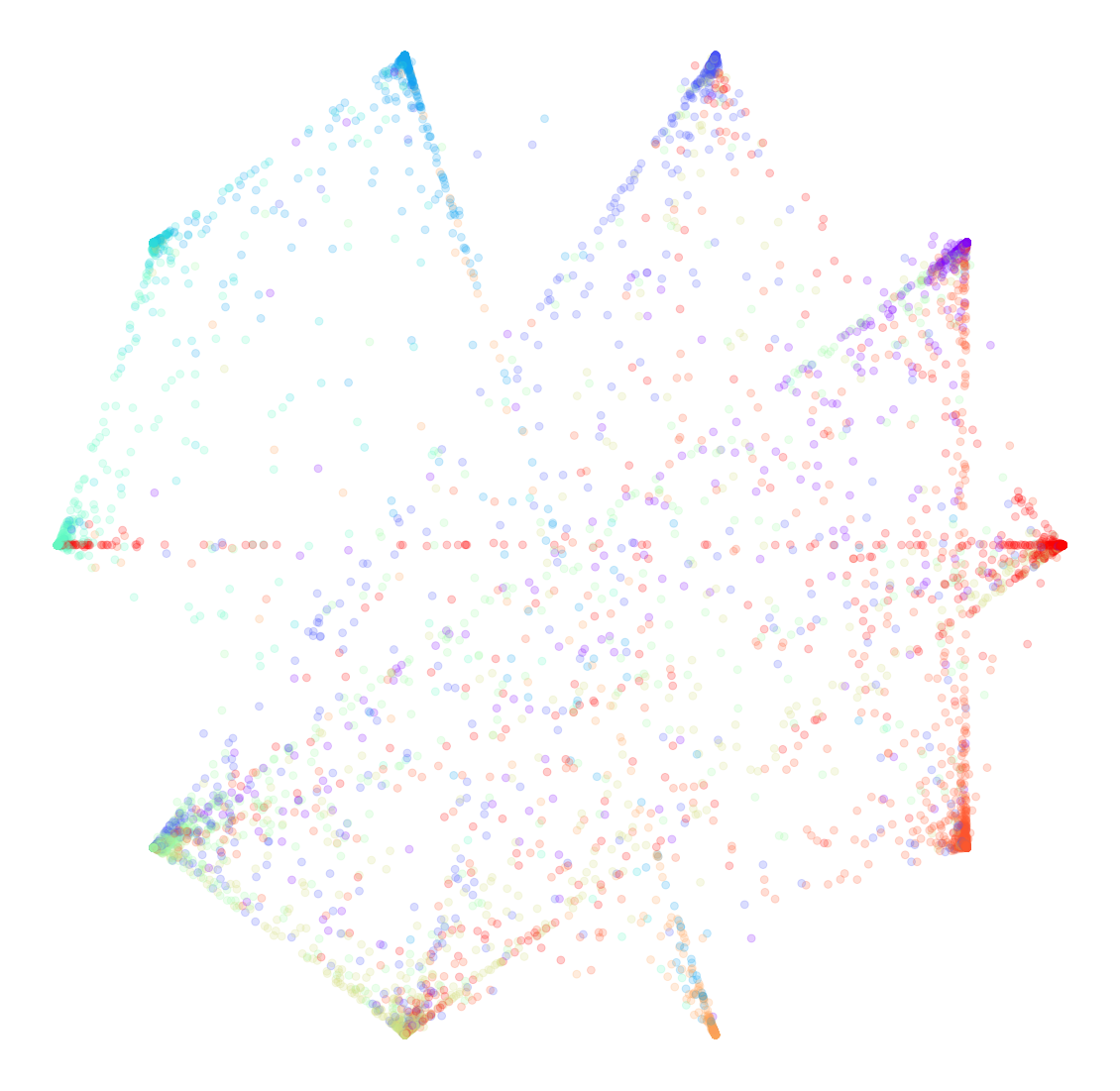}
      \caption{SCAN}
    \end{subfigure}
    \hspace{3mm}
    \begin{subfigure}[b]{0.22\textwidth}
      \includegraphics[width=\textwidth]{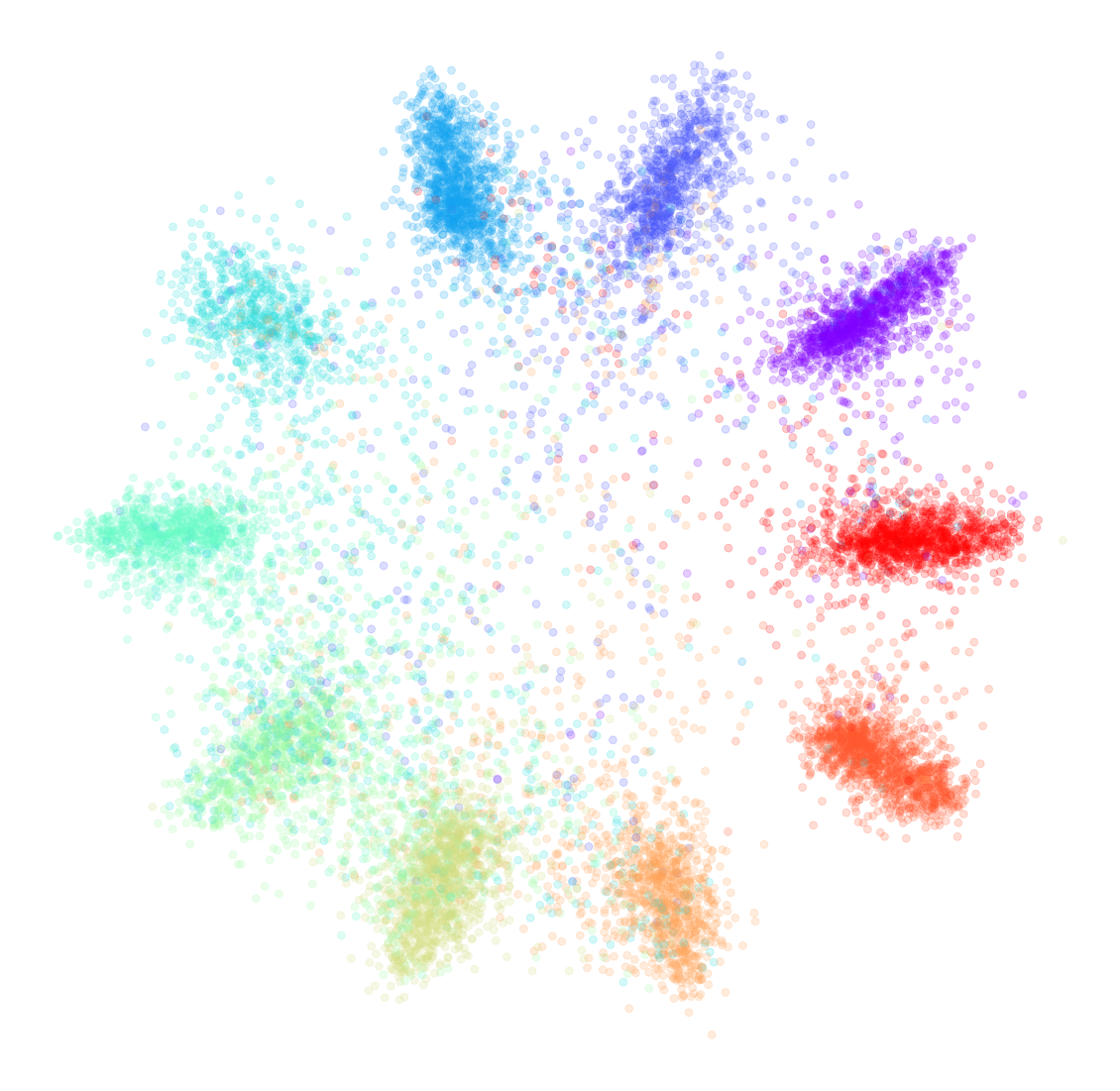}
      \caption{SCAN+\model~(epoch 50)}
    \end{subfigure}
    \hspace{3mm}
    \begin{subfigure}[b]{0.22\textwidth}
      \includegraphics[width=\textwidth]{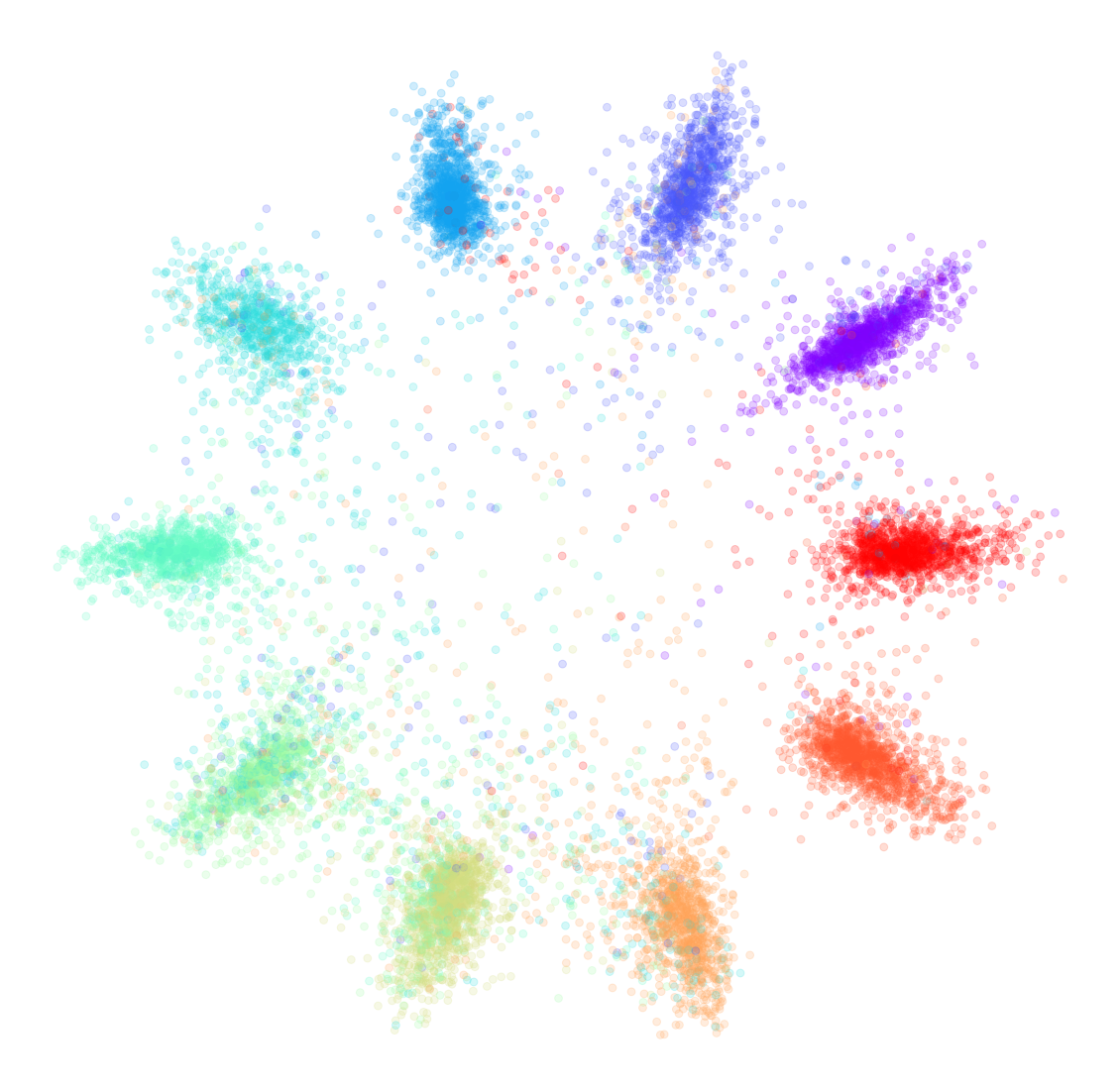}
      \caption{SCAN+\model~(epoch 100)}
    \end{subfigure}
    \hspace{3mm}
    \begin{subfigure}[b]{0.22\textwidth}
      \includegraphics[width=\textwidth]{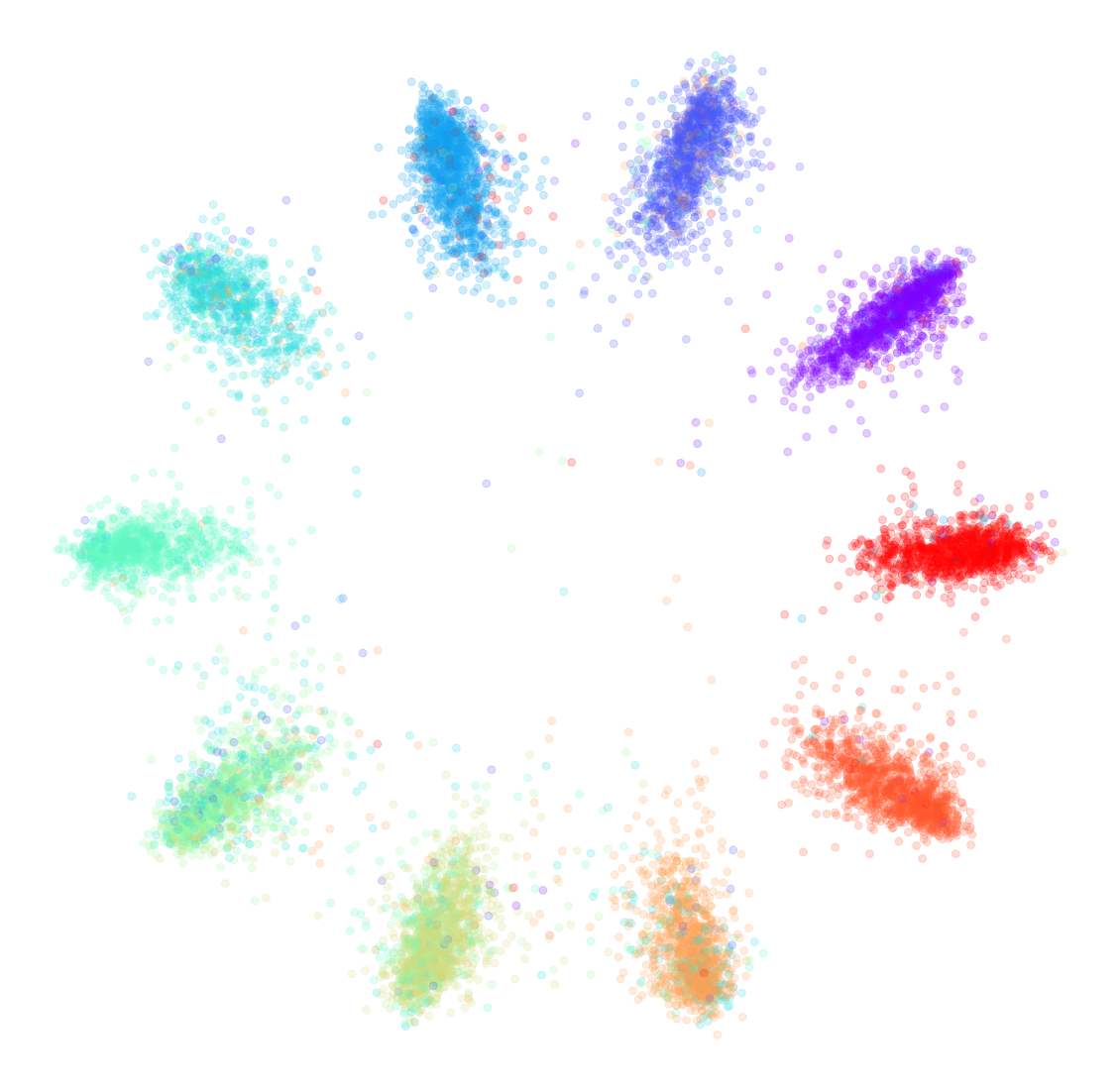}
      \caption{SCAN+\model~(epoch 200)}
    \end{subfigure}    
\caption{Visualization of clustering results on STL-10. The leftmost figure shows the results from the SCAN, while the right three figures display the intermediate results of our model on top of SCAN at different training epochs. This result demonstrates that \model{} effectively alleviates the overconfident predictions while enhancing the clustering quality.}
\label{fig:progress}
\end{minipage}
\end{figure*}

To evaluate the model's efficacy, we conduct an ablation study by repeatedly assessing its performance after removing each component. We also evaluate the accuracy of different selection and refurbishing strategies based on precision, recall, and the F1 score.  
\cutparagraphup
\paragraph{Ablation study.} The proposed model utilizes two robust learning techniques to cope with unclean samples: co-training and label smoothing. We remove each component from the full model to assess its efficacy. Table~\ref{Tab: Ablation} shows the classification accuracy of each ablation on the STL-10 dataset. \model{} with all components performs the best, implying that dropping any component results in performance degradation. We also compare a variant, which drops both label smoothing and co-training (i.e., MixMatch only). The effect of co-training is not evident in Table~\ref{Tab: Ablation}. Nevertheless, it improved the performance from 36.3\% to 39.6\% for the lowest noise ratio CIFAR-20 pseudo-labels when we set the base model as TSUC. This finding may suggest that co-training is more effective for pseudo-labels with high noise ratios. The co-training structure showed additional stability in training. Due to space limitation, we report these findings in the supplementary material.

\begin{table}[t!]
\centering
\scalebox{0.9}{
\begin{tabular}{l|ccc}
\toprule
Dataset              & CIFAR-10 & CIFAR-20 & STL-10  \\ \midrule
SCAN  & 88.7 & 50.6 & 81.4  \\ 
SCAN  + Gupta~\emph{et al.} & 88.3 / 89.5  & 53.2 / 53.3 & 84.2 / 84.3 \\ 
SCAN + DivideMix & 86.5 / 87.9 & 53.5 / 53.6 & 80.6 / 83.9 \\ 
SCAN + M-correction & 81.6 / 88.6 & 48.5 /50.4 &  81.1 / 81.3  \\ 
SCAN + P-correction & 87.7 / 88.7 & 49.5 / 50.5  & 81.2 /81.4 \\ \midrule
SCAN + \model{} (ours) &   \textbf{90.1}/ 90.1   & \textbf{54.3} / 54.5  &  \textbf{86.6} / 86.7\\\bottomrule
\end{tabular}}
\caption{Performance comparison with other possible add-on modules (Last / Best accuracy)}  
\label{Tab:robust_learning}
\end{table}

\cutparagraphup
\paragraph{Comparison with other possible add-on modules} 
As an alternative of \model{}, one may combine the extant robust learning algorithms (e.g., M-correction~\cite{arazo2019unsupervised}, P-correction~\cite{yi2019probabilistic}, and DivideMix~\cite{li2020dividemix}) or another previously proposed add-on module (e.g., Gupta~\textit{et al.}~\cite{gupta2020unsupervised}) on top of SCAN~\cite{van2020scan}. Table~\ref{Tab:classification} summarizes the comparisons to four baselines. For a fair comparison, we employed SCAN as the initial clustering method and applied each baseline on top of it.\footnote{For methods employing an ensemble of clusterings (\emph{e.g.}, Gupta~\textit{et al.}), we employed multiple SCAN models with random initialization. We also applied the same semi-supervised method (\emph{i.e.}, MixMatch).} As shown in the results, improving noisy clustering is non-trivial as some baselines show even worse results after the refinement (\emph{e.g.}, DivideMix, M-correction, P-correction). While Gupta~\textit{et al.} effectively refines the results, its improvement is limited when the initial clustering is reasonably accurate (\emph{e.g.}, CIFAR-10). In contrast, \model{} achieves consistent improvements in all datasets with non-trivial margins, showing that a carefully designed robust learning strategy is critical to the performance gain.

\begin{figure}[t!]
\centering

\includegraphics[width=0.36\textwidth]{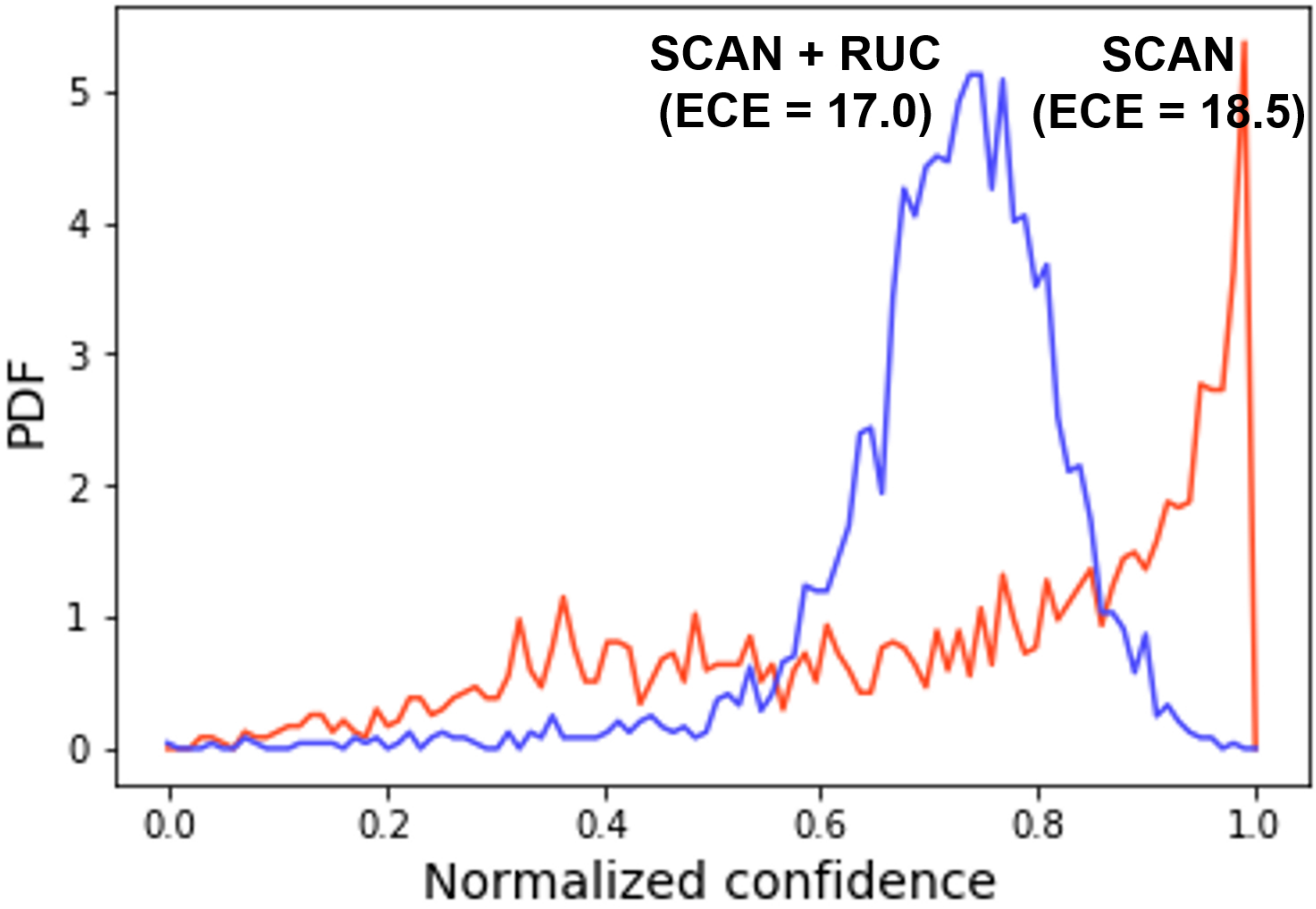}
    \caption{Confidence distribution for noise samples from the STL-10 dataset. \model{} shows more widely distributed confidence and produces better calibration.} 
    \label{fig:calibration}
\end{figure}

\begin{figure*}[t!]
\centering
\begin{minipage}{0.55\textwidth}
\captionsetup{width=.95\linewidth}
\includegraphics[width=0.99\columnwidth]{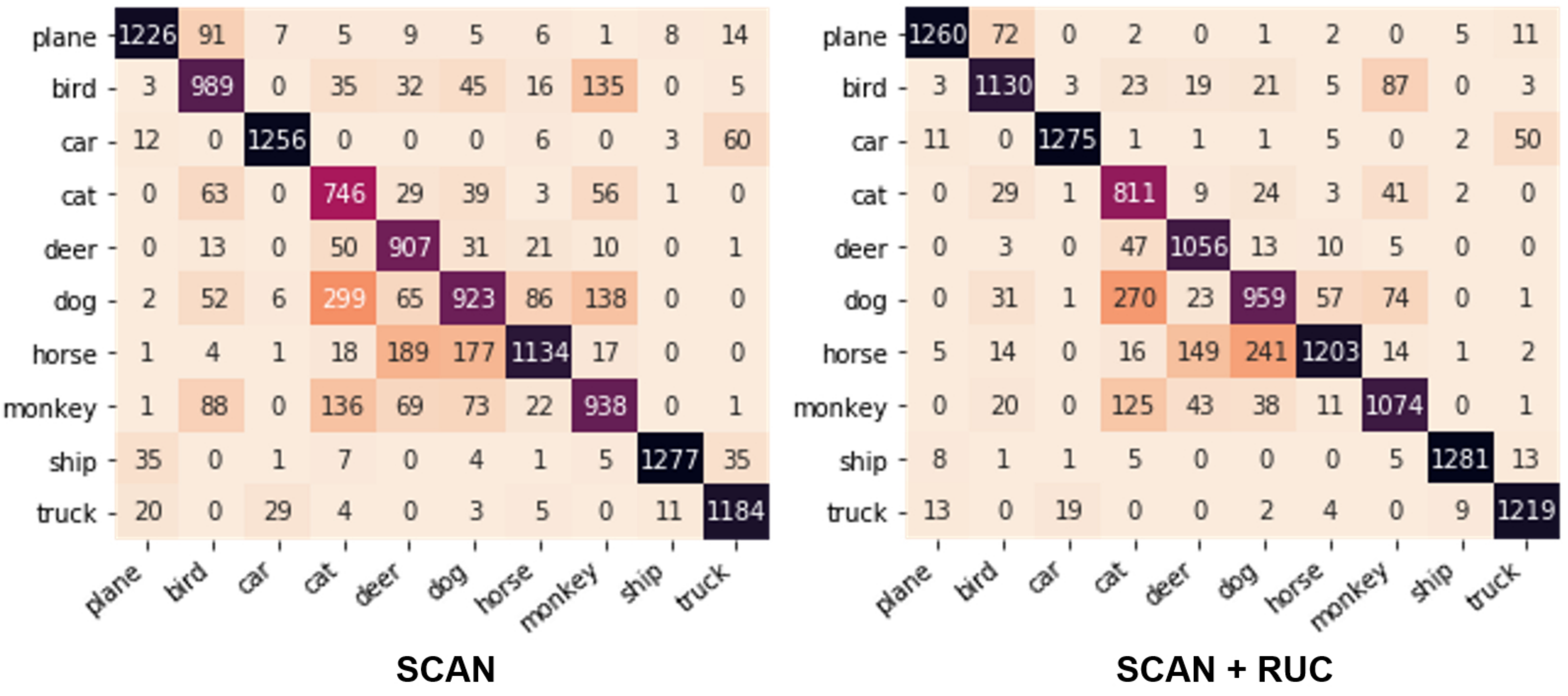}
    \caption{Confusion matrices of the SCAN and SCAN+\model~results on STL-10. The row names are predicted class labels, and the columns are the ground-truths.}
    \label{fig:confusion_matrix}
\end{minipage}
\hspace{0.18cm}
\begin{minipage}{0.41\textwidth}
\captionsetup{width=.9\columnwidth}
\includegraphics[width=0.95\columnwidth]{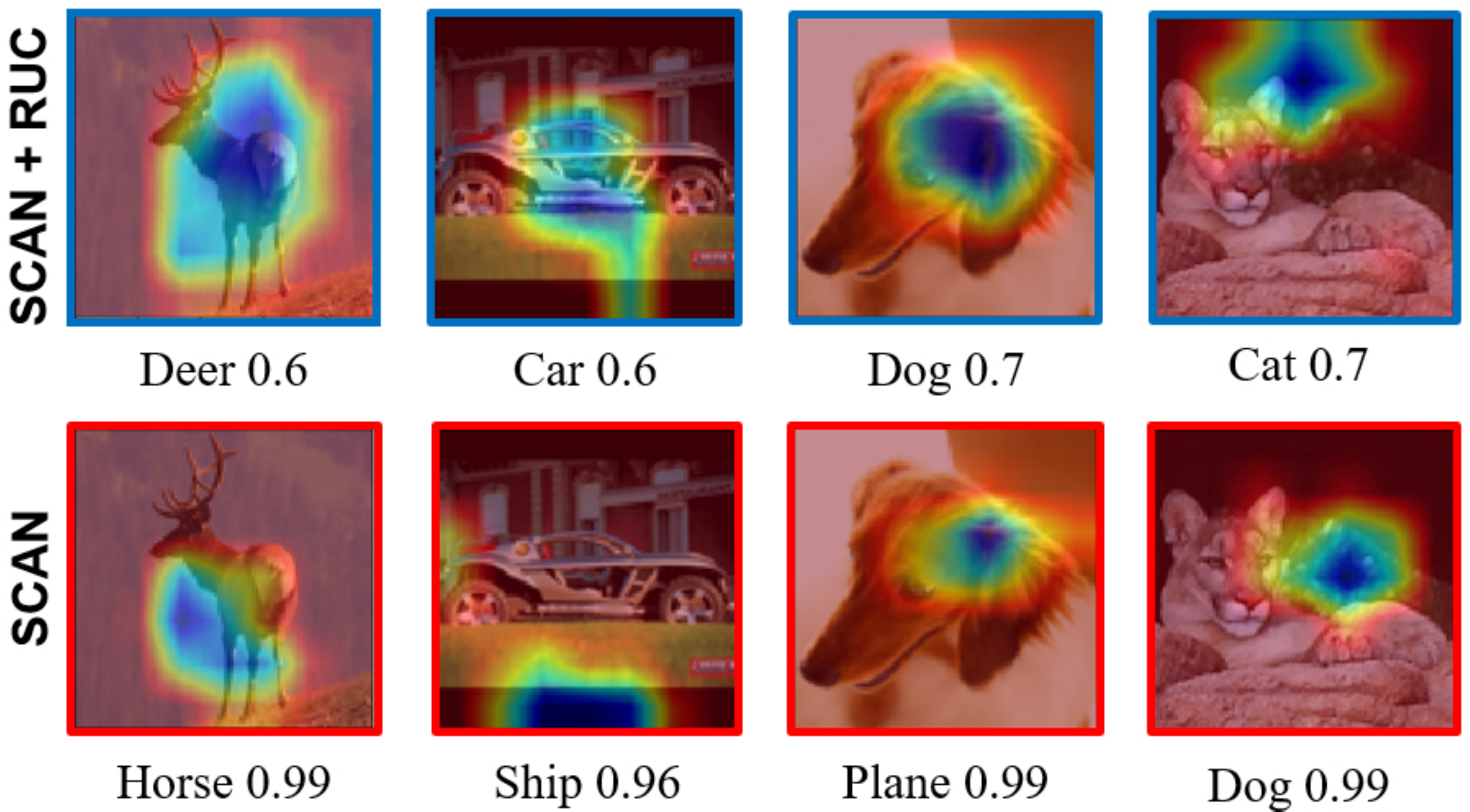}
    \caption{Class activation maps and the model's confidence on STL-10. The highlighted area indicates where the model focused to classify the image.
    }
    \label{fig:qualitive}
\end{minipage}
\vspace{-0.1in}
\end{figure*}

\subsection{In-Depth Performance Analysis} 
So far, we showed that \model{} improves existing baselines substantially, and its components contribute to the performance gain. We now examine \model{}'s calibration effect and present a qualitative analysis by applying it to the state-of-the-art base model, SCAN~\cite{van2020scan}. We will also demonstrate the role of \model{} in handling adversarially crafted noise. \looseness=-1

\cutparagraphup
\paragraph{Confidence calibration.} Many unsupervised clustering algorithms are subject to overconfident results due to their entropy-based balancing~\cite{hanmitigating,van2020scan}. If a model is overconfident to noisy samples, separating the clean and unclean set becomes challenging, and this can induce overall performance degradation. Figure~\ref{fig:calibration} shows the calibration effect of ~\model. SCAN's confidence is highly concentrated near 1, while our model's confidence is widely distributed over [0.6, 0.8]. 
We also report the degree of calibration quality using Expected Calibration Error (ECE)~\cite{guo2017calibration}:
\begin{equation}
\text{ECE} = \sum_{m=1}^M \frac{|B_m|}{n} |acc(B_m) - conf(B_m)|,
\end{equation} where $n$ is the number of data, $B_m$ is the $m$-th group from equally spaced buckets based on the model confidence over the data points; $acc(B_m)$ and $conf(B_m)$ are the average accuracy and confidence over $B_m$. Lower ECE of \model \space in Figure~\ref{fig:calibration} implies that our approach led to better calibrations. 

To observe this effect more clearly, we visualize the clustering confidence result at different training epochs in Figure~\ref{fig:progress}.  Unlike the result of SCAN in which the overly clustered sample and the uncertain sample are mixed, the result of SCAN+\model{} shows that the sample's class distribution has become smoother, and uncertain samples disappeared quickly as training continues.

\cutparagraphup
\paragraph{Qualitative analysis.} 
We conducted a qualitative analysis to examine how well \model{}~corrects the initial misclassification in pseudo-labels. Figure~\ref{fig:confusion_matrix} compares the confusion matrices of  SCAN and the SCAN+\model{} for STL-10. A high concentration of items on the diagonal line confirms the advanced correction effect of \model{} for every class. Figure~\ref{fig:qualitive} compares how the two models interpreted class traits based on the Grad-CAM~\cite{selvaraju2017grad} visualization on example images. The proposed model shows a more sophisticated prediction for similar images.

\cutparagraphup
\paragraph{Robustness to adversarial noise.}
Clustering algorithms like SCAN introduce pseudo-labels to train the model via the Empirical Risk Minimization (ERM) method~\cite{vapnik2013nature}. ERM is a learning principle that minimizes the averaged error over the sampled training data (i.e., empirical risk) to find the model with a small population risk (i.e., true risk). However, ERM is known to be vulnerable to adversarial examples, which are crafted by adding visually imperceptible perturbations to the input images~\cite{madry2018towards,zhang2017mixup}. 

\begin{figure}[!h]
    \centering
    \begin{subfigure}[b]{.495\linewidth}
        \centering\captionsetup{width=1.0\linewidth}
        \includegraphics[width=\linewidth]{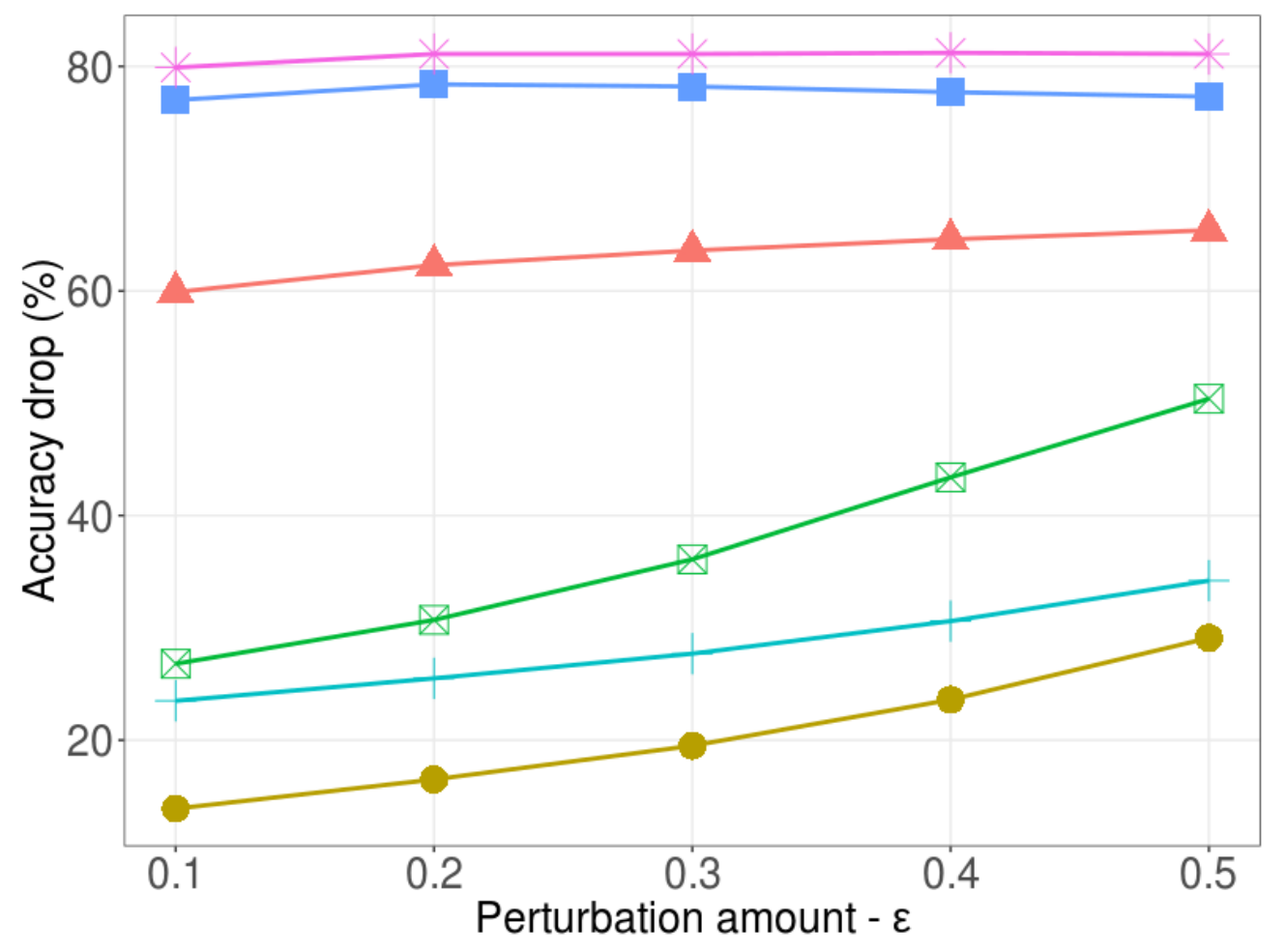}
        \caption{FGSM attack}
    \end{subfigure}
    \begin{subfigure}[b]{.495\linewidth}
        \centering\captionsetup{width=1.0\linewidth}
        \includegraphics[width=\linewidth]{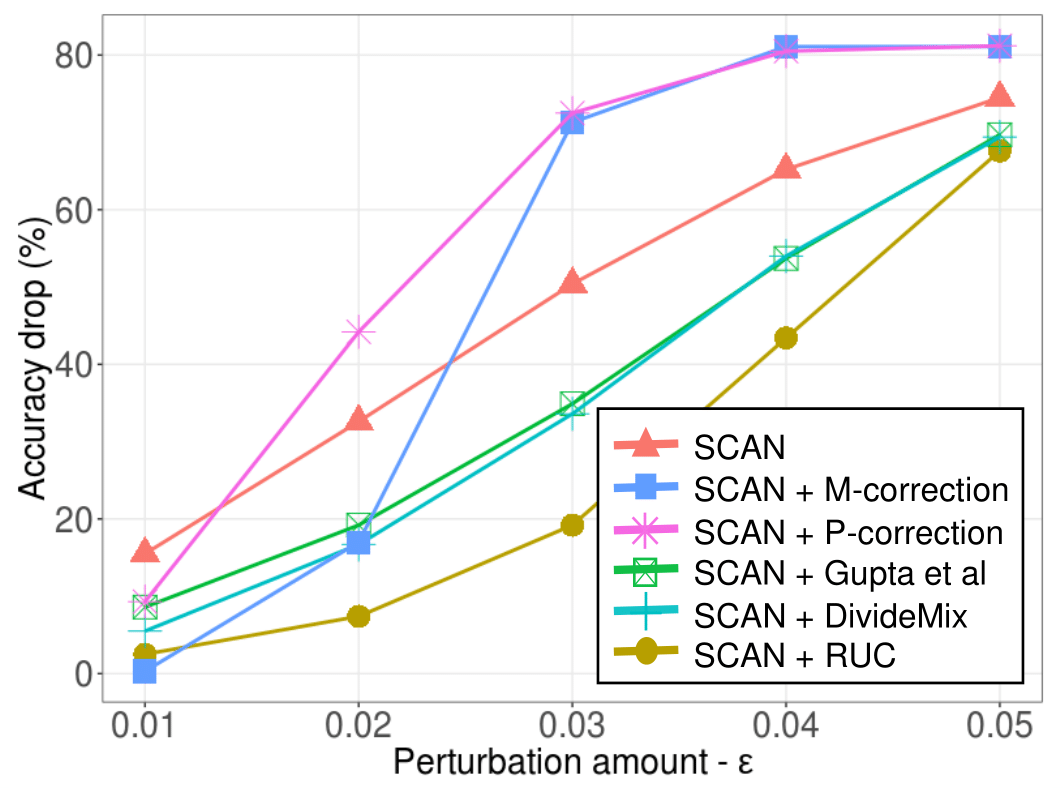}
        \caption{BIM attack}
    \end{subfigure}
    \caption{\model{}'s robustness to adversarial attacks, experimented with different perturbation rate $\epsilon$. Robust learning components are important in handling the adversarial noise.}
    \label{fig:adversarial-attack}
\end{figure}

Here, we show that \model{} improves robustness against the adversarial noise. We conduct an experiment on STL-10 using adversarial perturbations of FGSM~\cite{goodfellow2014explaining} and BIM~\cite{kurakin2016adversarial} attacks, whose directions are aligned with the gradient of the loss surface of given samples. Figure~\ref{fig:adversarial-attack} compares the model's ability to handle the adversarial noise. Models based on MixMatch (Gupta~\textit{et al.}, DivideMix, \model{}) outperform the rest, probably because the calibration effect of MixUp prevents overconfident predictions. Among them, \model{} achieves superior improvement, demonstrating that robust learning components, such as careful filtering, label smoothing, and co-training, can also handle the adversarial noise (see the supplementary material for further details).

\vspace{-4mm}
\section{Conclusion}
This study presented \model{}, an add-on approach for improving existing unsupervised image clustering models via robust learning. Retraining via robust training helps avoid overconfidence and produces more calibrated clustering results. As a result, our approach achieved a meaningful gain on top of two state-of-the-art clustering methods. Finally, \model{} helps the clustering models to be robust against adversarial attacks. We expect robust learning will be a critical building block to advance real-world clustering solutions. 

\small{\paragraph{Acknowledgements.}
This work was supported in part by the IBS (IBS-R029-C2) and the Basic Science Research Program through the NRF funded by the Ministry of Science and ICT in Korea (No. NRF-2017R1E1A1A01076400),
Institute of Information \& communications Technology Planning \& Evaluation (IITP) grant  (2020-0-00153 and 2016-0-00464), Samsung Electronics, HPC support funded by MSIT \& NIPA.
}

\nocite{papernot2016distillation,warde201611} 
\newpage
\balance


{\small
\bibliographystyle{ieee_fullname}
\bibliography{egbib}
}

\clearpage
\appendix
\section{Supplementary Material}



\subsection{Release}
Codes, training details, and the downloadable link for trained models are available at~\url{https://github.com/deu30303/RUC}.

\subsection{Training Details}
Our model employed the ResNet18~\cite{he2016deep} architecture following other baselines~\cite{hanmitigating,ji2019invariant,van2020scan}. Before retraining, note that we randomly initialize the final fully connected layer and replace the backbone network with a newly pretrained one from an unsupervised embedding learning algorithm as done in SimCLR~\cite{chen2020simple}. This random re-initialization process helps avoid the model from falling into the same local optimum.
The initial confidence threshold $\tau_{1}$ was set as 0.99, and the number of neighbors $k$ to divide the clean and noise samples was set to 100. 
The threshold $\tau_{2}$ for refurbishing started from 0.9 and increased by 0.02 in every 40 epochs. The label smoothing parameter $\epsilon$ was set to 0.5.
The initial learning rate was set as 0.01, which decays smoothly by cosine annealing. The model was trained for 200 epochs using SGD with a momentum of 0.9, a weight decay of 0.0005. The batch size was 100 for STL-10 and 200 for CIFAR-10 and CIFAR-20. We chose $\lambda_{u}$ as {25, 50, 100} for CIFAR-10, STL-10 and CIFAR-20. The $w_{b}$ value was calculated by applying min-max normalization to the confidence value of the counter network $f_{\theta^{(c)}}$.
Random crop and horizontal flip were used as a weak augmentation, which does not deform images' original forms.  RandAugment~\cite{cubuk2020randaugment} was used as a strong augmentation. We report all transformation operations for strong augmentation strategies in Table~\ref{Tab:Strong_aug}. The number of transformations and magnitude for all the transformations in RandAugment was set to 2.  

\begin{table}[!h]
\centering
\scalebox{0.9}{
\begin{tabular}{l|cc}
\toprule
Transformation & Parameter & Range \\ \midrule
AutoContrast    &  - &  -\\
Equalize    &  - & -\\
Identity    &  - & -\\
Brightness  &  $B$ & [0.01, 0.99]\\
Color&   $C$ & [0.01, 0.99]\\
Contrast    &  $C$ & [0.01, 0.99]\\
Posterize    &  $B$ & [1, 8]\\
Rotate    &  $\theta$ & [-45, 45]\\
Sharpness & $S$ & [0.01, 0.99]\\
Shear X, Y &  $R$ & [-0.3, 0.3]\\
Solarize &  $T$ & [0, 256]\\
Translate X, Y &  $\lambda$ & [-0.3, 0.3]\\
\bottomrule 
\end{tabular}}
\caption{List of transformations used in RandAugment}
\label{Tab:Strong_aug}
\end{table}

To evaluate class assignment, the Hungarian method~\cite{kuhn1955hungarian} was used to map the best bijection permutation between the predictions and ground-truth. We also note that the computational cost of \model{} is not a huge burden. It took less than 12 hours to run 200 epochs with 4 TITAN Xp processors for all datasets. \looseness = -1

\subsection{Sampling strategy analysis.}

\begin{figure}[h!]
\centering
\includegraphics[width=0.33\textwidth]{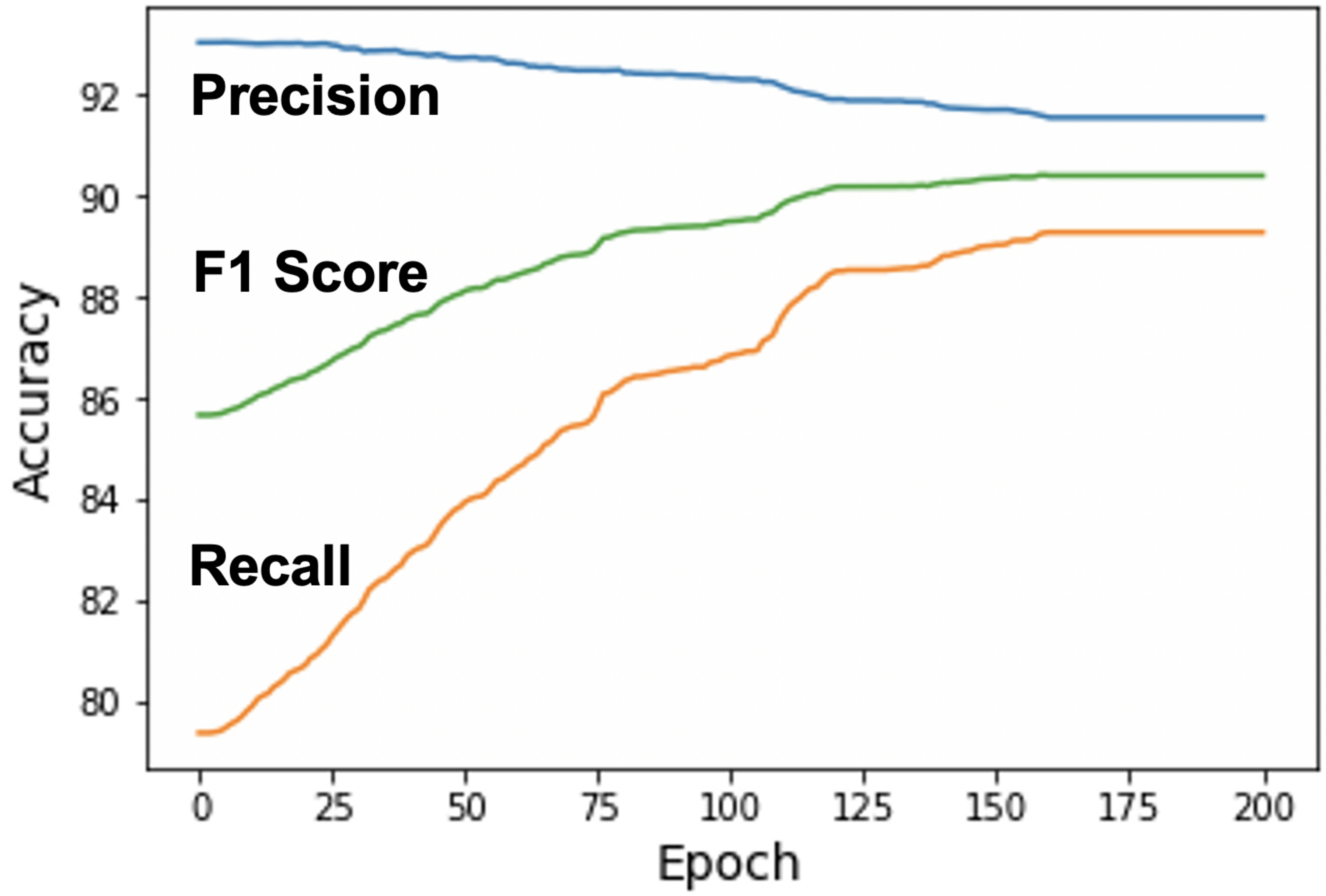}
\caption{Changes of sampling accuracy across each epoch on our model} 
\label{fig:f1score} 
\end{figure}

We evaluate the quality of the clean set generated from three sampling strategies (See Table~\ref{tab:strategy}). Overall, precision was the highest for the hybrid strategy, whereas recall was the highest for the metric-based strategy. We also tested the co-refurbish accuracy over the epochs. Figure~\ref{fig:f1score} displays the change of precision, recall, and the F1-score using confidence-based sampling on the STL-10 dataset. The model's precision drops slightly as the number of epochs increases, but the recall increases significantly. The F1-score, which shows the overall sampling accuracy of the clean set, increased about 5\% over 200 epochs. It can be interpreted a higher rate of true-positive cases than the false-positive cases in the refurbished samples, which means that the model could successfully correct the misclassified unclean samples. Overall, we find the current hybrid selection strategy can distinguish clean sets relatively well since the selected samples benefit from both strategies' merits. This strategy, however, cannot always achieve the best performance. Further development of the selection strategy will help increase the proposed \model{} model.

\begin{table}[!ht]
\centering
\scalebox{0.75}{
\begin{tabular}{l|ccc|ccc|ccc}
\toprule
\multicolumn{1}{l|}{\multirow{3}{*}{Strategy}} & \multicolumn{3}{c}{CIFAR-10}                      & \multicolumn{3}{c}{CIFAR-20}                      & \multicolumn{3}{c}{STL-10}                        \\  \cmidrule{2-4} \cmidrule{5-7} \cmidrule{8-10}
\multicolumn{1}{l|}{}                          & \multicolumn{1}{c}{C} & \multicolumn{1}{c}{M} & \multicolumn{1}{c}{H} & \multicolumn{1}{c}{C} & \multicolumn{1}{c}{M} & \multicolumn{1}{c}{H} & \multicolumn{1}{c}{C} & \multicolumn{1}{c}{M} &\multicolumn{1}{c}{H}\\ \midrule
Precision       &          92.6               &    91.6             &    93.7  &          59.5                 &    59.0           &     63.6     &          93.0               &        87.6       &   94.2       \\
Recall       &          93.5               &    93.2           &     89.3   &          83.1                 &    88.5          &    77.4       &          79.4               &        94.4                & 78.3 \\
F1 Score      &      93.0   &         92.4      &      91.4    &      69.3     &              70.8    &    69.8   &         85.7                &          90.9     &   85.5       \\ \bottomrule
\end{tabular}
}
\caption{Quality of the clean set (C : Confidence, M : Metric, H : Hybrid)}
\label{tab:strategy}
\end{table}

\subsection{Hyper-parameters of Sampling Strategies}
\begin{figure}[!h]
    \centering
    \begin{subfigure}[b]{.495\linewidth}
        \centering\captionsetup{width=1.0\linewidth}
        \includegraphics[width=\linewidth]{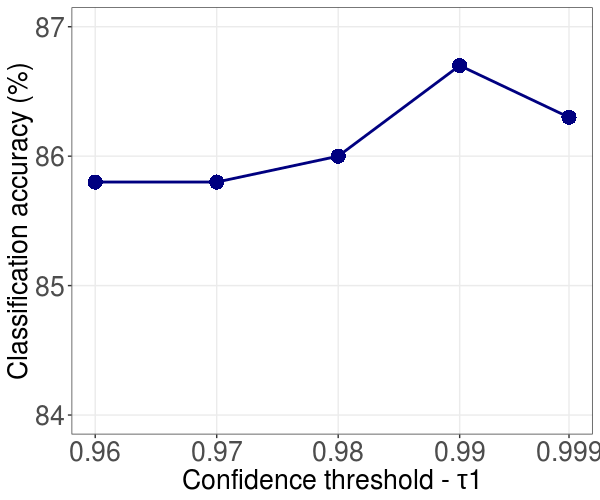}
        \caption{Effect of $\tau_1$}
    \end{subfigure}
    \begin{subfigure}[b]{.495\linewidth}
        \centering\captionsetup{width=1.0\linewidth}
        \includegraphics[width=\linewidth]{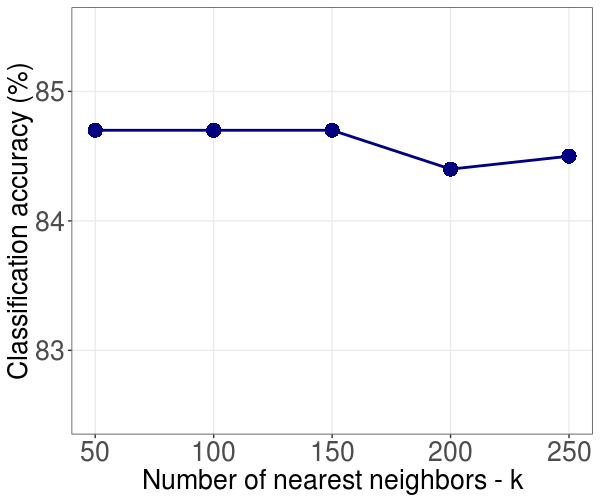}
        \caption{Effect of $k$}
    \end{subfigure}
    \caption{Analysis of the accuracy on the STL-10 dataset across two hyper-parameters: (a) $\tau_1$ from the confidence-based strategy and (b) $k$ from the metric-based strategy.}
    \label{fig:hyperanalysis}
\end{figure}

We investigate the effect of hyper-parameters from two sampling strategies: $\tau_1$ and $k$. $\tau_1$ is the threshold for selecting clean samples in the confidence-based strategy, and $k$ is the number of neighbors for the kNN classifier in the metric-based strategy. Figure~\ref{fig:hyperanalysis} summarizes the effect of each hyper-parameter. In the case of $\tau_1$, the final accuracy reaches the highest at $\tau_1 = 0.99$ and starts to decrease. Small $\tau_1$ extracts clean samples with higher recall and lower precision, while large $\tau_1$ extracts clean samples with higher precision and lower recall. Hence, balancing between the precision and recall through appropriate $\tau_1$ can lead to better performance. Meanwhile, the number of nearest neighbors $k$ does not significantly affect the final accuracy. Given $k$ within the reasonable range, our model consistently produces results of high performance. 

For remaining hyper-parameters ($\lambda_{u}$, $\epsilon$, $\tau_{2}$), our model resorted to a standard hyper-parameter setting that is commonly used in practice. For example, we set $\lambda_U$ following earlier works~\cite{berthelot2019mixmatch,li2020dividemix}, choose $\epsilon=0.5$ as the mean of $\text{Uniform}(0,1)$, and choose $\tau_{2}$ to be a reasonably high value, similar to $\tau_{1}$. Empirically, we find the model is oblivious to these parameters (see Table~\ref{Tab:hyperparameter}).  

\begin{table}[!h]
\centering
    \begin{minipage}{1.5in}
        \scalebox{0.7}{
        \begin{tabular}{c|ccc}
        \toprule
        $\lambda_{u}$ & 25 & 50 & 100 \\ \midrule
        STL-10 & 86.20 & 86.7 & 85.74 \\
        CIFAR-10 & 90.3 & 90.07 & 89.21 \\
        CIFAR-20 & 53.00 & 53.05 & 53.50 \\\bottomrule
        \end{tabular}}
    \end{minipage}
    \hspace{1mm}
    \begin{minipage}{1.5in}
        \scalebox{0.7}{
        \begin{tabular}{c|ccc}
        \toprule
        $\epsilon$ & 0.4 & 0.5 & 0.6 \\ \midrule
        CIFAR-10 & 90.31 & 90.3 & 90.27 \\ \toprule
        $\tau_{2}$ & 0.85 & 0.9 & 0.95  \\ \midrule
        CIFAR-10 & 90.28 & 90.3 & 90.28 \\ \bottomrule
        \end{tabular}}
    \end{minipage}
\caption{Hyper-parameter analyses ($\lambda_{u}$, $\epsilon$, $\tau_{2}$)}
\label{Tab:hyperparameter}
\end{table}

\subsection{Additional Analysis for RUC on TSUC}
\begin{figure}[h!]
\centering
\captionsetup{width=.9\linewidth}
\includegraphics[width=0.6\columnwidth]{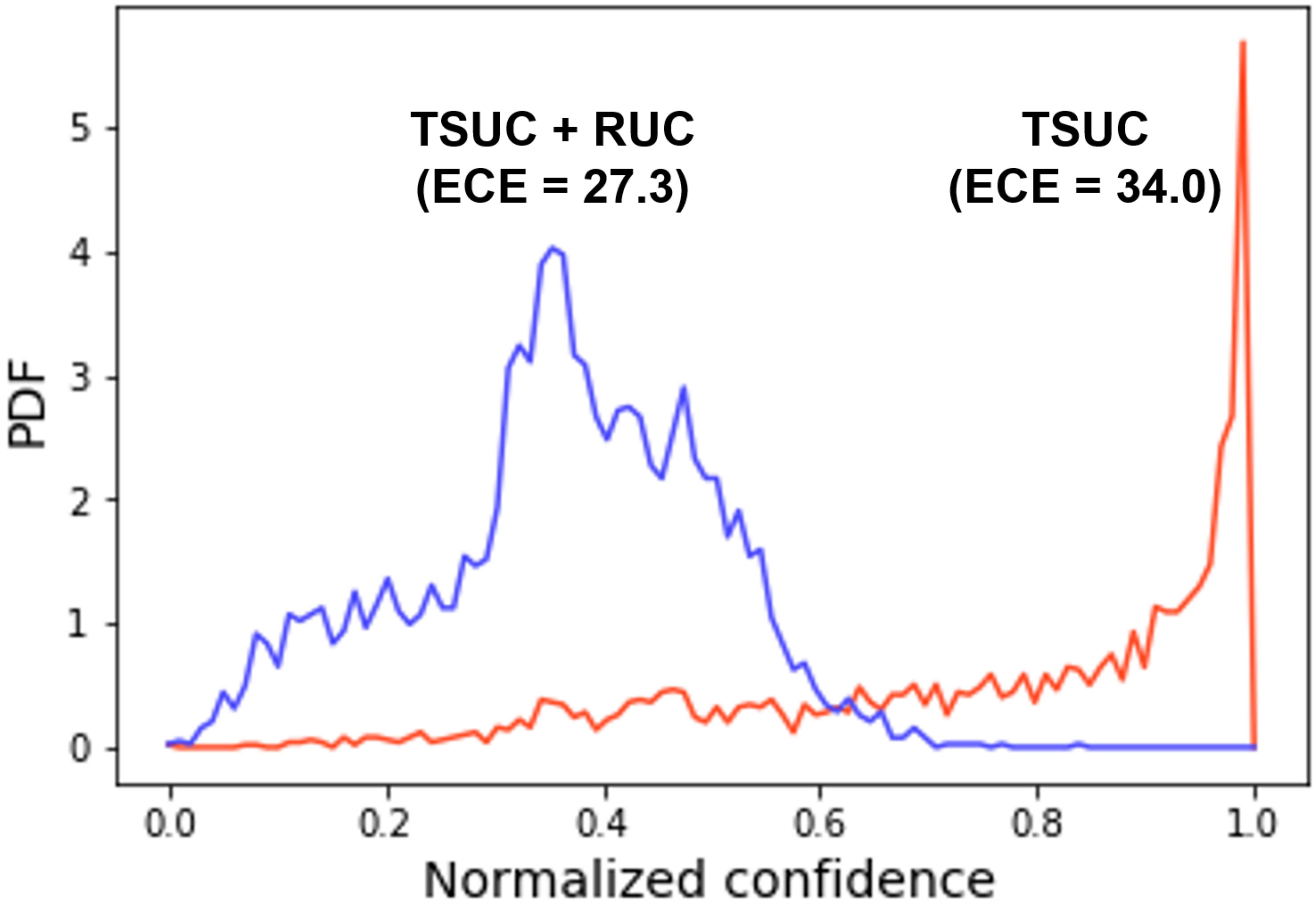}
    \caption{Confidence distribution for noise sample on STL-10 with the base model TSUC.}
    \label{fig:tsuc_conf}
\end{figure}

Many recent unsupervised clustering algorithms are subject to overconfident results because of their entropy-based balancing~\cite{hanmitigating,van2020scan}. If a model is overconfident to noisy samples, separating the clean set and the unclean set becomes challenging, which can induce the overall performance degradation. We evaluated the calibration effect of \model \space on top of TSUC in Figure~\ref{fig:tsuc_conf}. TSUC's confidence is highly concentrated near 1, while our model's confidence is widely distributed. We also report the degree of calibration quality using the Expected Calibration Error (ECE)~\cite{guo2017calibration}:

\begin{equation}
\text{ECE} = \sum_{m=1}^M \frac{|B_m|}{n} |acc(B_m) - conf(B_m)|,
\end{equation} where $n$ is the number of data points, $B_m$ is the $m$-th group from equally spaced buckets based on the model confidence over the data points; $acc(B_m)$ and $conf(B_m)$ are the average accuracy and confidence over $B_m$ respectively. TSUC's high ECE implies that TSUC is more overconfident than SCAN.  
Lower ECE of TSUC + \model{} case in Figure~\ref{fig:tsuc_conf} implies that our add-on process led to better calibrations. 

We also evaluated quality of the clean set from TSUC under three sampling strategies. The results are shown in Table~\ref{tab:tsuc_strategy}. Overall, the precision is the highest for the hybrid strategy, whereas the recall is the highest for the metric-based strategy, as same as the SCAN's results. Meanwhile, confidence-based strategies in TSUC showed low precision, which implies that TSUC is not well-calibrated and highly overconfident. 

\begin{table}[!h]
\centering
\scalebox{0.77}{
\begin{tabular}{l|ccc|ccc|ccc}
\toprule
\multicolumn{1}{l|}{\multirow{3}{*}{Strategy}} & \multicolumn{3}{c}{CIFAR-10}                      & \multicolumn{3}{c}{CIFAR-20}                      & \multicolumn{3}{c}{STL-10}                        \\  \cmidrule{2-4} \cmidrule{5-7} \cmidrule{8-10}
\multicolumn{1}{l|}{}                          & \multicolumn{1}{c}{C} & \multicolumn{1}{c}{M} & \multicolumn{1}{c}{H} & \multicolumn{1}{c}{C} & \multicolumn{1}{c}{M} & \multicolumn{1}{c}{H} & \multicolumn{1}{c}{C} & \multicolumn{1}{c}{M} &\multicolumn{1}{c}{H}\\ \midrule
Precision       &          80.9               &    84.2             &    82.7  &          40.9                 &    41.8           &     43.0     &          68.2              &        69.0       &   71.4       \\
Recall       &          69.9               &    96.4           &     68.0   &          47.4                &    90.3          &    45.5       &          78.3               &        79.4
& 76.2 \\
F1 Score      &      69.5   &         89.9      &      74.6    &      43.9     &              85.5    &    44.2   &         72.9                &          73.8     &   73.7       \\ \bottomrule
\end{tabular}
}
\caption{Quality of the clean set regards to sampling strategies (C : Confidence, M : Metric, H : Hybrid)}
\label{tab:tsuc_strategy}
\end{table}

\subsection{Further Discussion on Co-Training}
\label{sup:co-training}
\begin{figure}[h!]
\centering
\captionsetup{width=.9\linewidth}
\includegraphics[width=0.6\columnwidth]{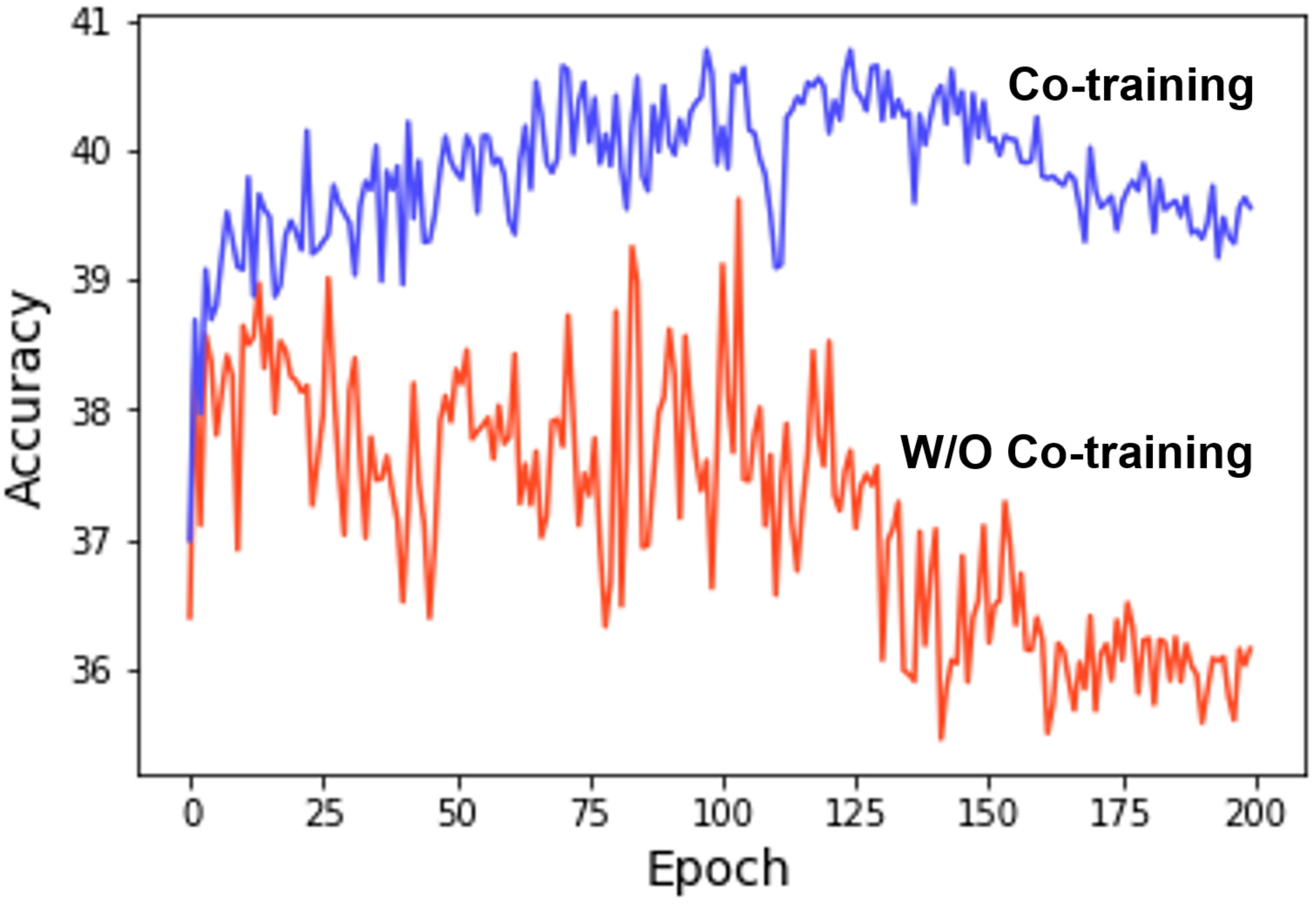}
    \caption{Changes of clustering accuracy across each epoch on CIFAR-20 with the base model TSUC}
    \label{fig:wo_co}
\end{figure}
Our model architecture introduces a co-training module where the two networks exchange their guesses for teaching each other via co-refinement. Due to the different learning abilities in two networks, disagreements from networks help filter out corrupted labels, which contributes to a substantial performance increase in unsupervised classification. Besides, the co-training structure provides extra stability in the training process.

Figure~\ref{fig:wo_co} compares our model and the same model without co-training based on classification accuracy across the training epoch. The model without co-training shows large fluctuations in accuracy; in contrast, the full model's accuracy remains stable and consistent throughout epochs. We speculate this extra stability comes from our model's ensemble architecture and the effect of loss correction. Corrected labels via ensemble predictions bring additional label smoothing. Therefore, it may reduce the negative training signals from unclean samples, which can lead to abrupt updates on the model parameters.

\subsection{Further Details on Adversarial Robustness}
Empirical risk minimization (ERM), a learning principle which aims to minimize the averaged error over the sampled training data (i.e., empirical risk), has shown remarkable success in finding models with small population risk (i.e., true risk) in the supervised setting~\cite{vapnik2013nature}. However, ERM-based training is also known to lead the model to memorize the entire training data and often does not guarantee to be robust on adversarial noise~\cite{madry2018towards,zhang2017mixup}. This weakness can also be inherited from several unsupervised clustering algorithms that introduce the ERM principle with their pseudo-labels, like SCAN~\cite{van2020scan}.

Adding \model{} to the existing clustering models improves robustness against adversarial noise. To demonstrate this, we conducted an experiment using adversarial perturbations of the FGSM~\cite{goodfellow2014explaining} and BIM~\cite{kurakin2016adversarial} attacks, whose directions are aligned with the gradient of the loss surface of given samples. The details of each attack are as follows:

\cutparagraphup
\paragraph{Fast Gradient Sign Method (FGSM)}
FGSM crafts adversarial perturbations by calculating the gradients of the loss function $J(\theta, \mathbf{x}, \mathbf{y})$ with respect to the input variables. The input image is perturbed by magnitude $\epsilon$ with the direction aligned with the computed gradients (Eq.~\eqref{eq:fgsm_equation}).
\begin{align}
    \mathbf{x}^{adv} = \mathbf{x} + \epsilon\cdot \text{sgn}(\nabla_{\mathbf{x}} J(\theta, \mathbf{x}, \mathbf{y})) \label{eq:fgsm_equation}
\end{align}

\cutparagraphup
\paragraph{Basic Iterative Method (BIM)}
BIM is an iterative version of FGSM attack, which generates FGSM based adversarial noise with small $\epsilon$ and applies the noise many times in a recursive way (Eq.~\eqref{eq:bim_equation}).
\begin{align}
    \hspace{-2mm}\mathbf{x}_0^{adv} &= \mathbf{x} \\
    \hspace{-2mm}\mathbf{x}_i^{adv} &= \text{clip}_{\mathbf{x}, \epsilon}( \mathbf{x}_{i-1}^{adv} + \epsilon\cdot \text{sgn}(\nabla_{\mathbf{x}_{i-1}^{adv}} J(\theta, \mathbf{x}_{i-1}^{adv}, \mathbf{y}))) \label{eq:bim_equation}
\end{align}
Clip function maintains the magnitude of noise below $\epsilon$ by clipping. For BIM attack experiments, we use five iterations with an equal step size. \smallskip

\begin{figure*}[t!]
\centering
\includegraphics[width=2.0\columnwidth]{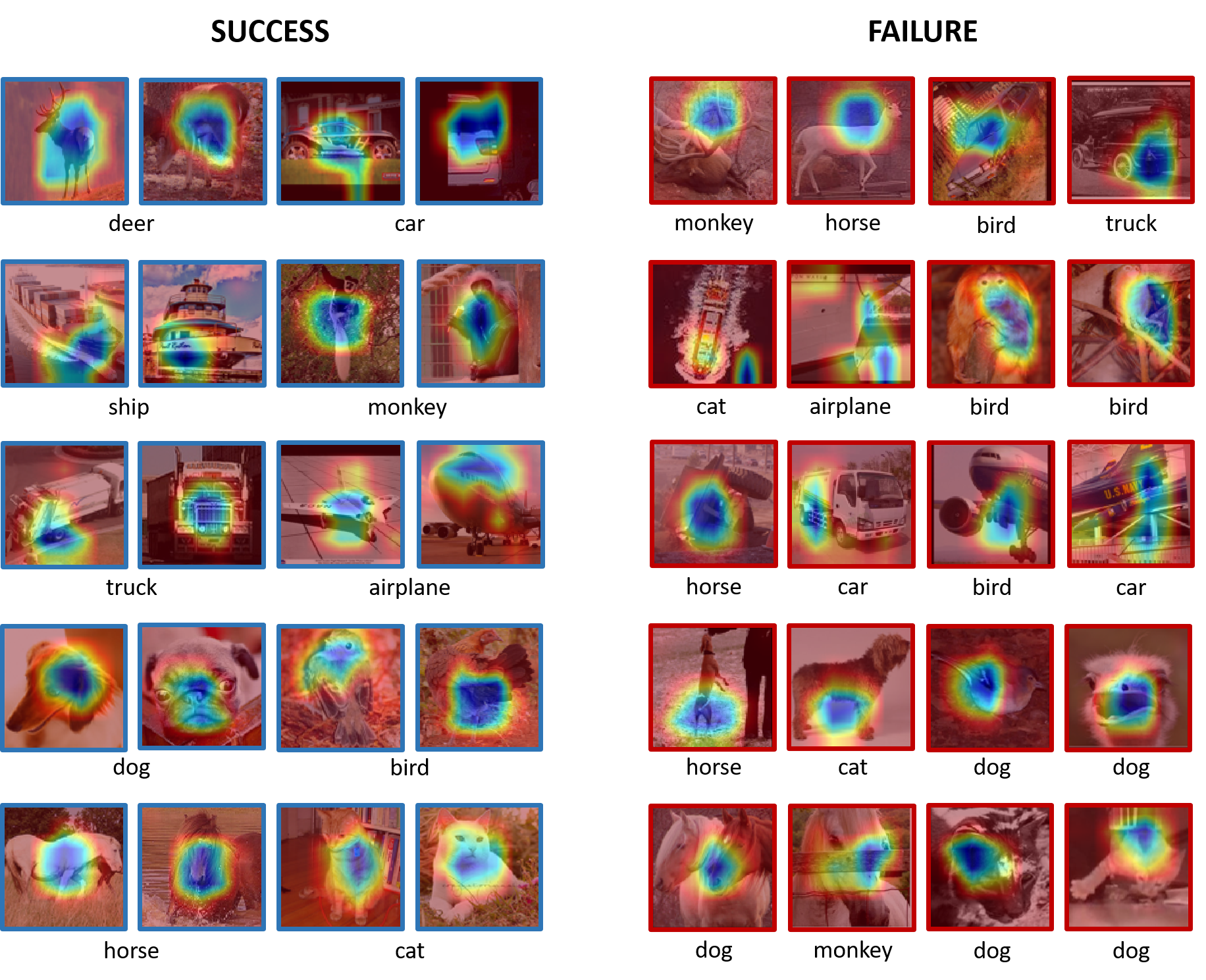}
\caption{Additional example of successes and failures from STL-10 where the highlighted part indicates how the model interprets class traits based on the Grad-CAM method (Blue frame: success case, Red frame: failure case).}
\label{fig:qualitative20}
\end{figure*}
Figure~\ref{fig:adversarial-attack} in our main manuscript compares the model's ability to handle adversarial attacks, which confirms that adding \model{} helps maintain the model accuracy better for both attack types. An investigation could guide us that this improved robustness is mainly due to the label smoothing techniques, which regularize the model to avoid overconfident results and reduce the amplitude of adversarial gradients with smoothed labels~\cite{papernot2016distillation,warde201611}.

\subsection{Additional Examples for Qualitative Analysis}
Figure~\textcolor{red}{\ref{fig:qualitative20}} shows additional examples for the visual interpretation from RUC on top of SCAN via the Grad-CAM algorithm~\cite{selvaraju2017grad}. Blue framed images are the randomly chosen success cases from STL-10, and the red-framed images are example failure cases. Overall, the network trained with our model can extract key features from the images. Even though the model sometimes fails, most of the failures occurred between visually similar classes (e.g., horse-deer, cat-dog, truck-car, dog-deer).

\end{document}